%% file: paper.tex
\documentclass{article}

\usepackage{graphicx}
\usepackage{epsfig,subfigure}
\usepackage{amsmath,amssymb}
\usepackage{float}
\usepackage{rotating}

\usepackage[round]{natbib}

\usepackage{hyperref}
\usepackage{varioref}

\usepackage{graphpap}
\begin{document}

\title{Towards Swarm Calculus:\\ Urn Models of Collective Decisions and Universal Properties of Swarm Performance}

\author{Heiko Hamann\\
  Department of Computer Science\\
  University of Paderborn\\
  Zukunftsmeile~1\\
  33102~Paderborn, Germany\\
  \it{heiko.hamann@uni-paderborn.de} 
}

\maketitle

\begin{abstract}
  Methods of general applicability are searched for in swarm
  intelligence with the aim of gaining new insights about natural
  swarms and to develop design methodologies for artificial swarms. An
  ideal solution could be a `swarm calculus' that allows to calculate
  key features of swarms such as expected swarm performance and
  robustness based on only a few parameters. To work towards this
  ideal, one needs to find methods and models with high degrees of
  generality. In this paper, we report two models that might be
  examples of exceptional generality. First, an abstract model is
  presented that describes swarm performance depending on swarm
  density based on the dichotomy between cooperation and
  interference. Typical swarm experiments are given as examples to
  show how the model fits to several different results. Second, we
  give an abstract model of collective decision making that is
  inspired by urn models. The effects of positive feedback probability,
  that is increasing over time in a decision making system, are
  understood by the help of a parameter that controls the feedback
  based on the swarm's current consensus. Several applicable methods,
  such as the description as Markov process, calculation of splitting
  probabilities, mean first passage times, and measurements of
  positive feedback, are discussed and applications to artificial and
  natural swarms are reported.

  %\keywords{swarm performance \and collective decision-making \and urn model \and positive feedback}
\end{abstract}

\section{Introduction}

The research of swarm intelligence is important in both biology to
gain new insights about natural swarms and also in fields dealing with
artificial swarms, such as swarm robotics, to obtain sophisticated
design methodologies. The ideal tool would allow to calculate
fundamental features of swarm behavior, such as performance,
stability, and robustness, and would need only a few observed
parameters in the case of natural swarm systems or a few designed
parameters in the case of artificial swarms. We call this highly
desired set of tools `swarm calculus' (calculus in its general
sense). The underlying idea is to create a set of mathematical tools
that are easily applied to a variety of settings in the field of swarm
intelligence. In addition, these tools should be easily combined which
would allow using them as mathematical building blocks for
modeling. Thus, models will surely be an important part of swarm
calculus.

General properties and generally applicable models need to be found to
obtain a general methodology of understanding and designing swarm
systems. Today it seems that only few models exist that have the
potential to become general swarm models. For example, swarm models in
biology are particularly distinguished by their
variety~\citep{okubo01,okubo86,vicsek10,edelstein06,camazine01}. Typically,
a specialized model is created for each biological challenge. It seems
that the desire for models with wide applicability to a collection of
natural swarms is rather low in that community. In the field of
artificial swarms, such as robot swarms, the desire for generality
seems to be bigger, which is, for example, expressed by several models
in swarm
robotics~\citep{hamann10_diss_springer,berman11,prorok11,milutinovic07,lerman05}. The
driving force for the creation of these models is to support the
design of swarm robotic systems within a maximal range of
applications. The focus of these models is on quantitative
features of the swarm behavior, such as the distribution of robots or
required times for certain tasks. However, there is a struggle between
the intended generality of the model and the creation of a direct
mapping between the model and the actual description of the individual
robot's behavior. A~higher degree of generality is achievable if the
demand for a detailed description of behavioral features is abandoned
and focus is set only on high-level features such as overall
performance or the macroscopic process of a collective decision. Such
high-level models can be expressed by concise mathematical
descriptions that, in turn, allow direct applications of standard
methods from statistics, linear algebra, and statistical mechanics. In
this paper\footnote{This paper is an extended version of
  \citet{hamann12b}. The main extensions are the method of deriving
  the probability of positive feedback based on observed decision
  revisions (Sec.~\ref{sec:modelOfCollDec}), a discussion of
  additional methodology such as Markov chains, splitting
  probabilities, and mean first passage times
  (Sec.~\ref{sec:methods}), and a comprehensive introduction of the
  Ehrenfest and the Eigen urn models.}, we present two models of
general properties of swarm systems concerning the dependence of swarm
performance on swarm density and the dependence of collective
decisions on positive feedback.

\section{Fundamentals of swarm performance and collective decisions}

In this section we define the concepts of swarm performance and
collective decision-making along with so-called urn models upon
which the collective-decision model is based.

\subsection{Swarm performance}
By `swarm performance' we denote the efficiency of a swarm concerning
a certain task. For example, the swarm performance can be a success
rate of how often the task is accomplished on average, it can be the
average speed of a swarm in collective motion etc. Here we are
interested in the swarm performance as a function over `swarm
density', which is how many agents are found on average within a
certain area. For the following reason, the function of swarm
performance depending on swarm density cannot simply be a linear
function. For a true swarm system, a very low density, which
corresponds situations with only a few agents in the whole area, has
to result in low performance because there is neither a lot of
cooperation between agents because they seldom meet nor a
significant speed-up. With increasing density, the performance
increases, on the one hand, because of a simple speed-up (e.g., two
robots clean the floor faster than one) and, on the other hand,
because of increasing opportunities of cooperation (assuming that
cooperation is an essential beneficial part of swarms). In natural
swarms such increases in performance with increasing swarm size are
revealed, for example, in productivity gains and also in the emergence
of increased division of labor as an indicator for increased
cooperation~\citep{jeanne96b,karsai98,gautrais02,jeanson07}. Even
superlinear performance increases are possible in this interval of
swarm density and was reported for a swarm of
robots~\citep{mondada05b}. For artificial swarms it was reported that
at some critical/optimal density~\citep{fontan96} the performance
curve will first level off and then decrease~\citep{arkin93} because
improvements in cooperation possibilities will be lower than the
drawback of high densities, namely interference~\citep{lerman02}. With
further increase of the density, the performance continues to decrease
as reported for multi-robot systems~\citep{goldberg97}. Hence, swarms
generally face a tradeoff between cooperation, which is beneficial,
and interference, which is usually obstructive, however, has sometimes
positive effects both within certain natural swarms and as a tool for
designing swarm algorithms~\citep{dussutour04,goldberg97}.

In the following we report that many swarm systems not only show
similar qualitative properties but show also similarities in the
actual shapes of their swarm performance over swarm size/density
graphs (see function of swarm performance in
Fig.~\vref{fig:perfAllInOne}). Examples are the performance of
foraging in a group of robots (Fig.~\vref{fig:fitted:lerman} and
Fig.~10a in~\citep{lerman02}), the activation dynamics and information
capacity in an abstract cellular automaton model of ants (Figs.~1b and
1c in~\citep{miramontes95}), and even in the sizes of social networks
(Fig. 8b in~\citep{strogatz01}). A~similar curve is also presented as
a hypothesis for per capita output in social wasps by
\citet{jeanne96b}. The existence of this general shape was already
reported by \citet{ostergaard01} in their expected performance model
for multi-robot systems in constrained environments:
\begin{quote}
  \emph{We know that by varying the implementation of a given task, we can
  move the point of ``maximum performance'' and we can change the
  shapes of the curve on either side of it, but we cannot change the
  general shape of the graph.}
\end{quote}
Traffic models of flow over density are related because traffic flow
can be increased when cars cooperatively share streets but the flow
decreases when the streets are too crowded and cars interfere too much
with each other. While the `fundamental diagram' of traffic
flow~\citep{lighthill55} is symmetric, more realistic models propose
at least two asymmetric phases of free and synchronized flow~(e.g.,
Fig 3(b) in~\citep{wong02}). Actual measurements on highways show
curves with shapes similar to Fig.~\ref{fig:perfAllInOne} (e.g., see
Fig.~6-4 in~\citep{mahmassani09}). In these models, there exist two
densities for a given flow (except for maximum flow) similar to the
situation here where we have two swarm densities for each swarm
performance (one smaller than the optimal density and one bigger than
the optimal density; the corresponding function that maps densities to
performance values is surjective, not bijective).

%% In the field of distributed and swarm robotics, it seems that there is
%% more literature about interference issues than about speed-ups by
%% increasing the robot group sizes or by cooperation.

\subsection{Collective decisions}

In the context of swarm intelligence, collective decision-making is a
process that is distributed over a group of agents without global
knowledge. Each agent decides based on locally sampled data such as
the current decision of its neighbors. There are many biological
systems showing collective decision-making, for example, food source
choice in honey bees~\citep{Seeley1990}, nest site selection in
ants~\citep{mallon01}, and escape route search in social
spiders~\citep{Saffre_Collective_Decision-making_1999}. Collective
decision-making systems are often modeled as positive-feedback systems
that utilize initial fluctuations which are amplified and that way
help to converge to a global
decision~\citep{deneubourg90,mallon01,nicolis11}. Interesting features
of collective decisions are, for example, the speed-accuracy
trade-off~\citep{nicolis11} or the influence of
noise~\citep{dussutour09,yates09}. Furthermore, it turns out that
positive feedback is not always productive but can also generate
irrational decisions~\citep{nicolis11}.

In the following, we limit our investigations to binary decision
processes because they allow for a concise mathematical notation and,
hence, allow a manageable application of mathematical methods. The
investigated systems are either inherently noisy (e.g., explicit
stochastic processes within the agents' behaviors) or can validly be
modeled as noisy processes (e.g., deterministic chaos with strong
dependence on the initial conditions). We are not interested in the
quality of the final decision---that is the utility of choosing
option~$A$ over option~$B$ or vice versa---and assume that there is no
initial bias to one or the other. In selecting an appropriate model
for collective decisions, our main concern is simplicity while keeping
focus also on the relation of how much is needed as input to the model
and how much is generated by it. We want to keep the number of
parameters small while achieving descriptions of qualitative aspects
of collective decisions as effects of the model. In order to obtain
these standards, we choose a minimal macroscopic model which has only
one state variable describing the current status of the collective
decision within the swarm (e.g., 80\% for option~$A$ and consequently
20\% for option~$B$).

For simplicity, we view the asynchronous distributed process of
collective decisions as a round-based game which allows only one agent
at a time to either revise its current decision or to convince a peer
to revise its current decision. The relation to natural systems is
that decision events are serialized and intermediate periods of time
are ignored. The influence of this assumption to the steady state
behavior is considered to be low. For this purpose, we re-interpret
well-known urn models as models of collective decisions and extend
them appropriately. We use simple models inspired by the urn model of
\citet{ehrenfest07} and by the urn model of \citet{eigen93}.

\subsubsection{Ehrenfest urn model}
This urn model was originally introduced by \citet{ehrenfest07} in the
context of dissipation, thermodynamics, statistical mechanics, and
entropy. The dynamics of the model is defined as follows. Our urn is
filled with $N$~marbles. Say, initially all marbles are blue. Whenever
we draw a blue one we replace it with a red marble. If we draw a red
marble we replace it with a blue one. Obviously, the two extreme
states of either having an urn full of blue marbles or an urn full of
red marbles are unstable. Similarly, this is true for all states of
unevenly distributed colors. To formalize this process, we keep track
of how the number of blue marbles (without loss of generality) changes
depending on how many blue marbles were in the urn at that time. We
can do this empirically or we can actually compute the average
expected `gain' in terms of blue marbles. For example, say at time~$t$
we have $B(t)=16$ blue marbles in the urn and a total of $N=64$
marbles. The probability of drawing a blue marble is therefore
$P_B=\frac{16}{64}=0.25$. The case of drawing a blue marble has to be
weighted by $-1$ because this is the change in terms of blue marbles
in that case. The probability of drawing a red marble is
$P_R=\frac{48}{64}=0.75$ which is weighted by $+1$. Hence, the
expected average change~$\Delta B$ of blue marbles per round depending
on the current number of blue marbles~$B=16$ is $\Delta
B(B=16)=0.25(-1)+0.75(+1)=0.5$. This can be done for all possible
states yielding
\begin{equation}
  \label{eq:ehrenfest:deltaB}
  \Delta B(B)=-2\cdot\frac{B}{64}+1 \enspace ,
\end{equation}
which is plotted in Fig.~\ref{fig:urnModels:ehrenfest}. Hence, the
average dynamics of this game is given by $B(t+1)=B(t)+\Delta
B(B(t))$.

The recurrence $B_t=B_{t-1}-2\frac{B_{t-1}}{64}+1$ can be solved by
generating functions~\citep{graham98}. For $B_0=0$ we
obtain the generating function
\begin{equation}
  G(z)= \sum_t\left(\sum_{k\le t}\left(\frac{62}{64}\right)^k\right)z^t \enspace .
\end{equation}
The $t$th coefficient $[z^t]$ of this power series is the closed form
for $B_t$. We get
\begin{equation}
  \label{eq:ehrenfestEvolutionAnalytically}
  [z^t]=B_t=\sum_{k\le t}\left(\frac{62}{64}\right)^k=\frac{1-\left(\frac{62}{64}\right)^t}{1-\frac{62}{64}} \enspace .
\end{equation}

Hence, for initializations $B_0=0$ and the symmetrical case $B_0=64$
the system converges in average rather fast to the equilibrium
$B=32$. The actual dynamics of this game is, of course, a stochastic
process which can, for example, be modeled by $B(t+1)=B(t)+\Delta
B(B(t)) + \xi(t)$, for a noise term~$\xi$.

Several generalizations have been proposed for this urn
model~\citep{krafft93,klein56}, however, these investigations mostly
focus on mathematically tractable variants. In the following we
report generalizations, which focus on applications to feedback
processes.

\setlength{\unitlength}{0.0035\textwidth}
\begin{figure}%[h!]
  \centering
  \subfigure[\label{fig:urnModels:ehrenfest}Ehrenfest urn model]{
    \begin{picture}(130,90)
      \put(2,95){\includegraphics[angle=270,width=130\unitlength]
        {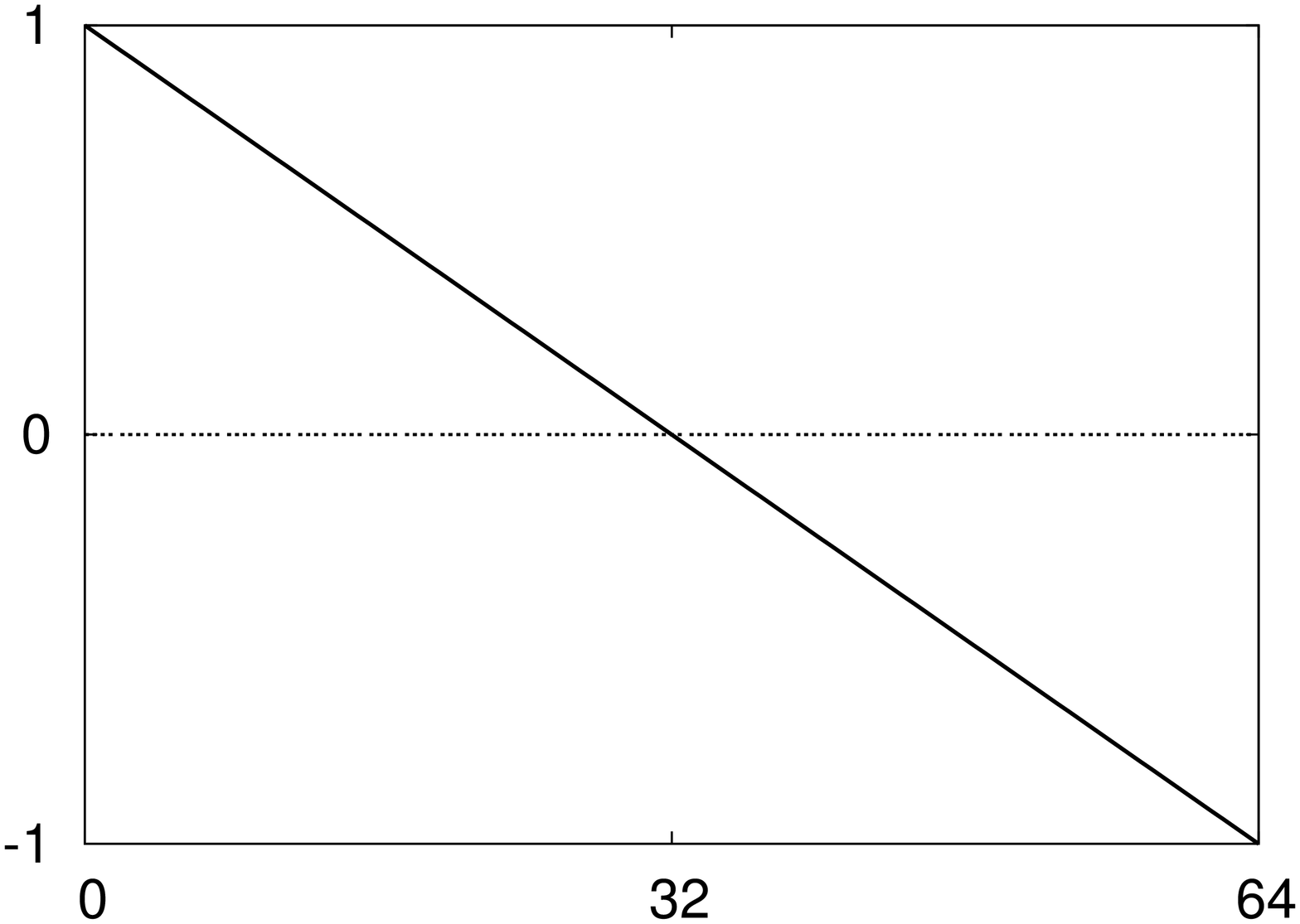}}
      \put(67,0){\small $B$}
      \put(0,45){\small\begin{sideways}$\Delta B$\end{sideways}}
      %@generator gnuplot
      %@script figs/data/collectiveDecision/urnModel/urnModels.plt
      %\graphpaper[10](0,0)(130,90)
    \end{picture}
  }
  \subfigure[\label{fig:urnModels:eigen}Eigen urn model (note $\Delta B(0)=\Delta B(64)=0$)]{
    \begin{picture}(130,90)
      \put(2,95){\includegraphics[angle=270,width=130\unitlength]
        {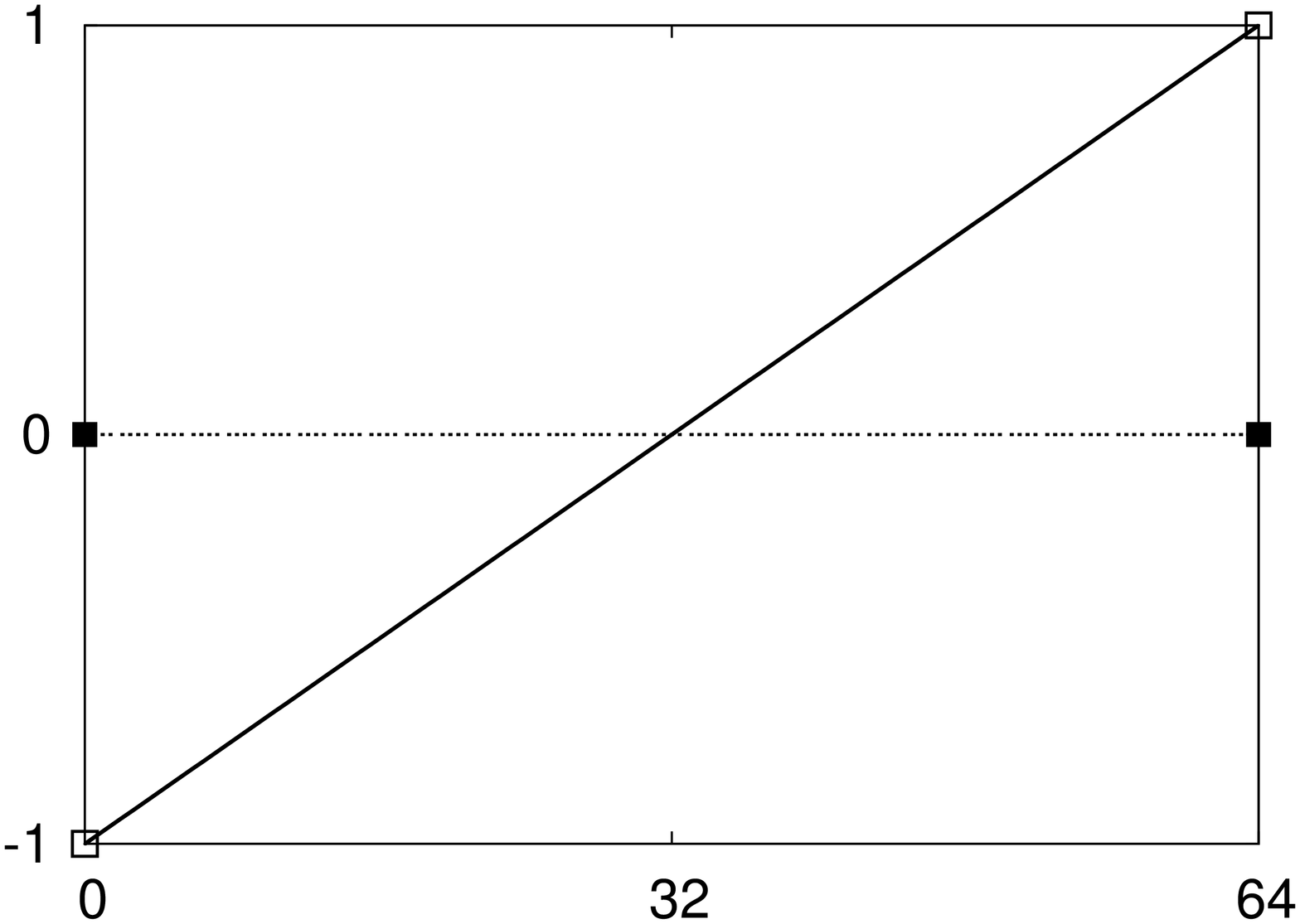}}
      \put(67,0){\small $B$}
      \put(0,45){\small\begin{sideways}$\Delta B$\end{sideways}}
      %@generator gnuplot
      %@script figs/data/collectiveDecision/urnModel/urnModels.plt
      %\graphpaper[10](0,0)(130,90)
    \end{picture}
  }
  \caption{\label{fig:urnModels}Expected average change in the urn
    models.}
\end{figure}

\subsubsection{Eigen urn model}
The Ehrenfest model can be interpreted as an example of a negative
feedback process. Deviations from the fixed point $B=32$ are corrected
by negative feedback (the predominant color will diminish on
average). \citet{eigen93} reported a similar urn model to show the
effect of positive feedback. In this model drawing a blue marble has
the effect of replacing a red marble by a blue one and vice versa. The
expected average change of blue marbles per round changes accordingly
to

\begin{equation}
  \label{eq:eigen:deltaB}
\Delta B(B)=
\begin{cases}
  2\frac{B}{64}-1, & \text{for } B\in[1,63]\\
  0, & \text{else}
\end{cases} \enspace .
\end{equation}
For a plot of $\Delta B(B)$ see Fig.~\ref{fig:urnModels:eigen}.

While we still have an expected change of $\Delta B=0$ for $B=32$ as
in the Ehrenfest model, the fixed point $B=32$ is unstable now as its
surrounding drives trajectories away from $B=32$ and towards the
stable fixed points $B=0$ and $B=64$ respectively. In
Sec.~\ref{sec:collectiveDecisions}, we introduce a more general urn
model that takes the intensity of positive feedback as a
parameter. This model can be used to investigate collective decisions
in swarms.

\section{Universal properties of swarm performance}

Having identified the two main components (cooperation and
interference) and the typical shape of these graphs, we can define a
simple model. The idea is to fit this model to empirical data for
verification and predictions.

\subsection{Simple model of swarm performance}

For a given bounded, constant area~$A$ the swarm density~$\rho$ is
defined by the swarm size~$N$ according to~$\rho=N/A$. Also a dynamic
area~$A(t)$ could be assumed but throughout this paper we want to keep
swarm density and swarm size interchangeable based on the
identity~$\rho=N/A$. Although for a given swarm density the swarm
performance might be quantitatively and qualitatively different for
different areas, here we focus on describing such swarm--performance
functions separately. We define the swarm performance~$\Pi$ depending on
swarm size~$N=\rho A$ by
\begin{equation}\label{eq:perf:fullEq}
  \Pi(N)=C(N)(I(N)-d)=a_1N^b a_2\exp(cN) \enspace ,
\end{equation}
for parameters $c<0$ (decreasing exponential function), $a_1,a_2>0$
(scaling), $b>0$, and~$d\ge 0$ (see
Fig.~\ref{fig:perfAllInOne}). Parameter~$d$ is subtracted to force a
decrease to zero ($\lim_{N\rightarrow \infty}I(N)-d=0$). The swarm
performance depends on two components, $C$ and~$I$. First, the swarm
effort without negative feedback is defined by the cooperation
function (see also Fig.~\ref{fig:perfAllInOne})
\begin{equation}\label{eq:perf:cooperation}
  C(N)=a_1N^b \enspace .
\end{equation}
This function can be interpreted as the potential for cooperation in a
swarm that would exist without certain constraints, such as physical
collisions or other spatial restrictions. The same formula was used
by~\citet{breder54} to model the cohesiveness of a fish school and by
\citet{bjerknes10} to model swarm velocity in emergent taxis. However,
they used parameters of~$b<1$ while we are also using values of
$b>1$. In principle this is a major difference because $b<1$
represents a sublinear performance increase due to cooperation and
$b>1$ represents a superlinear increase. Whether such a direct
interpretation of resulting parameter settings is instrumental is
unclear. Especially when analyzing the product of both cooperation and
interference functions (Eq.~\ref{eq:perf:fullEq}) it is seen that the
steepness depends on contributions from both functions which can be
differing considerably, for example, based on scaling. Second, the
interference function (see also Fig.~\ref{fig:perfAllInOne}) is
defined by
\begin{equation}\label{eq:perf:interference}
  I(N)=a_2\exp(cN)+d \enspace ,
\end{equation}
with $d$ used for scaling (e.g., $\lim_{N\rightarrow
  \infty}I(N)=d$). The interference function can also be interpreted
as the swarm performance achievable without cooperation, that is,
achievable swarm performance without positive feedback. The
exponential decrease of the interference function seems to be a
reasonable choice because, for example, the Ringelmann effect
according to \citet{ingham74} implies also a nonlinear decrease of
individual performance with increasing group size; see also
\citep[p. 236]{kennedy01}.
%efficiency of hashing over load factor~\citep{Knuth1998}
Nonlinear effects that decrease efficiency with increasing swarm size
are plausible due to negative feedback processes such as the collision
avoidance behavior of one robot triggers the collision avoidance
behavior of several others in high density situations.  Still, there
are many options of available nonlinear functions but best results
were obtained with exponential functions. Also Fig.~10b
in~\citep{lerman02} shows an exponentially decreasing efficiency per
robot in a foraging task.

\subsection{Examples}

To prove the wide applicability of this simple model we fit it to some
swarm performance plots that are available. We investigate four
scenarios: foraging in a group of robots~\citep{lerman02}, collective
decision making~\citep{hamann10a} based on
BEECLUST~\citep{schmickl_BEECLUST_2010}, aggregations in tree-like
structures and reduction to shortest paths~\citep{hamann06} similar
to~\citep{hamann08a}, and the emergent taxis scenario (also sometimes
called `alpha algorithm', \citet{nembrini02,bjerknes07}).

Given the data of the overall performance, the four parameters $a_1$,
$a_2$, $b$, $c$ of~$\Pi(N)$ (Eq.~\ref{eq:perf:fullEq}) can be directly
fitted. This was done for the three examples shown in
Figs.~\ref{fig:fitted:lerman}, \ref{fig:fitted:symBreakScan}, and
\ref{fig:fitted:diplomarbeit}. The equation can be well fitted to the
empirical data (for details about the fitted functions see the
appendix). In case of the foraging scenario
(Fig.~\ref{fig:fitted:lerman}) we also have data of the efficiency per
robot. We can use the model parameters~$a_2$ and~$c$, that were
obtained by fitting the model to the overall performance, to predict
the efficiency per robot, which is a function that we suppose to be
proportional to effects by interference. This is done by scaling the
interference function~$I(N)$ linearly and plotting it against the
efficiency per robot. The result is shown in
Fig.~\ref{fig:fitted:lerman}.

\setlength{\unitlength}{0.004\textwidth}
\begin{figure}
  \centering
  \subfigure[\label{fig:perfAllInOne}Model of swarm performance~$\Pi$ based on cooperation and interference, examples of swarm performance~$\Pi$ (Eq.~\ref{eq:perf:fullEq}), cooperation (Eq.~\ref{eq:perf:cooperation}), and interference (Eq.~\ref{eq:perf:interference}) depending on swarm size; $a_1 = 0.002$, $a_2=1$, $b = 2.5$, $c = -0.12$.]{
    \begin{picture}(110,85)
      \put(-3,88){\includegraphics[angle=270,width=120\unitlength]
        {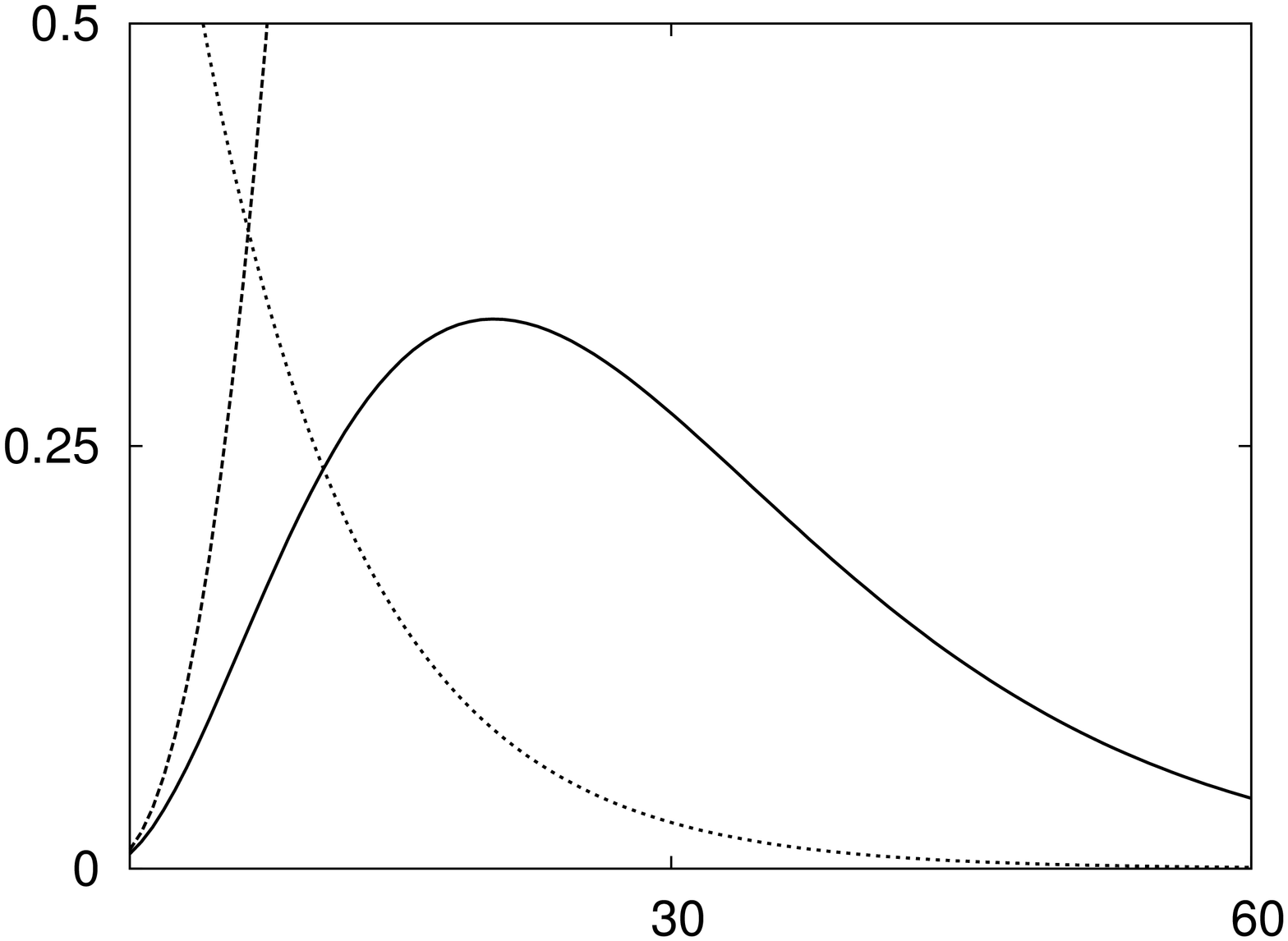}}
      \put(-3,46){$\Pi$}
      \put(58,0){$N$}
      % @generator gnuplot
      % @script figs/data/performance/allInOne.plt
      %\graphpaper[10](0,0)(110,85)
      \put(40,60){\scriptsize swarm performance $\Pi$}
      \put(29,77){\scriptsize cooperation}
      \put(27,15){\scriptsize interference}
    \end{picture}
  }
  \hspace{1.5mm}
  \subfigure[\label{fig:fitted:lerman}Foraging in a group of robots, Eq.~\ref{eq:perf:fullEq} fitted to group efficiency (upper solid line), prediction of interference (lower solid line, efficiency per robot, linearly rescaled), data points extracted from Fig.~10 in Lerman et al.~\citep{lerman02}; $\Pi$~gives group/robot efficiency.]{
    \begin{picture}(120,85)
    \put(-3,88){\includegraphics[angle=270,width=120\unitlength]
        {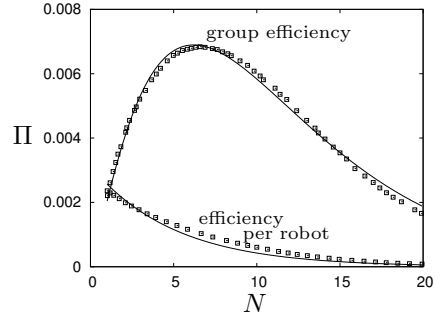}}
      \put(-3,46){$\Pi$}
      \put(60,0){$N$}
      \put(48,24){\scriptsize efficiency}
      \put(60,20){\scriptsize per robot}
      \put(50,75){\scriptsize group efficiency}
      %@generator gnuplot
      %@script figs/data/performance/lerman/lermanDiagram.plt
      %@data figs/data/performance/lerman/lermanDiagramData
      %\graphpaper[10](0,0)(110,85)
    \end{picture}
  }
  \subfigure[\label{fig:fitted:symBreakScan}Collective decision
      making~\citep{hamann10a} based on
      BEECLUST~\citep{schmickl_BEECLUST_2010}.]{
    \begin{picture}(110,85)
    \put(-3,88){\includegraphics[angle=270,width=120\unitlength]
        {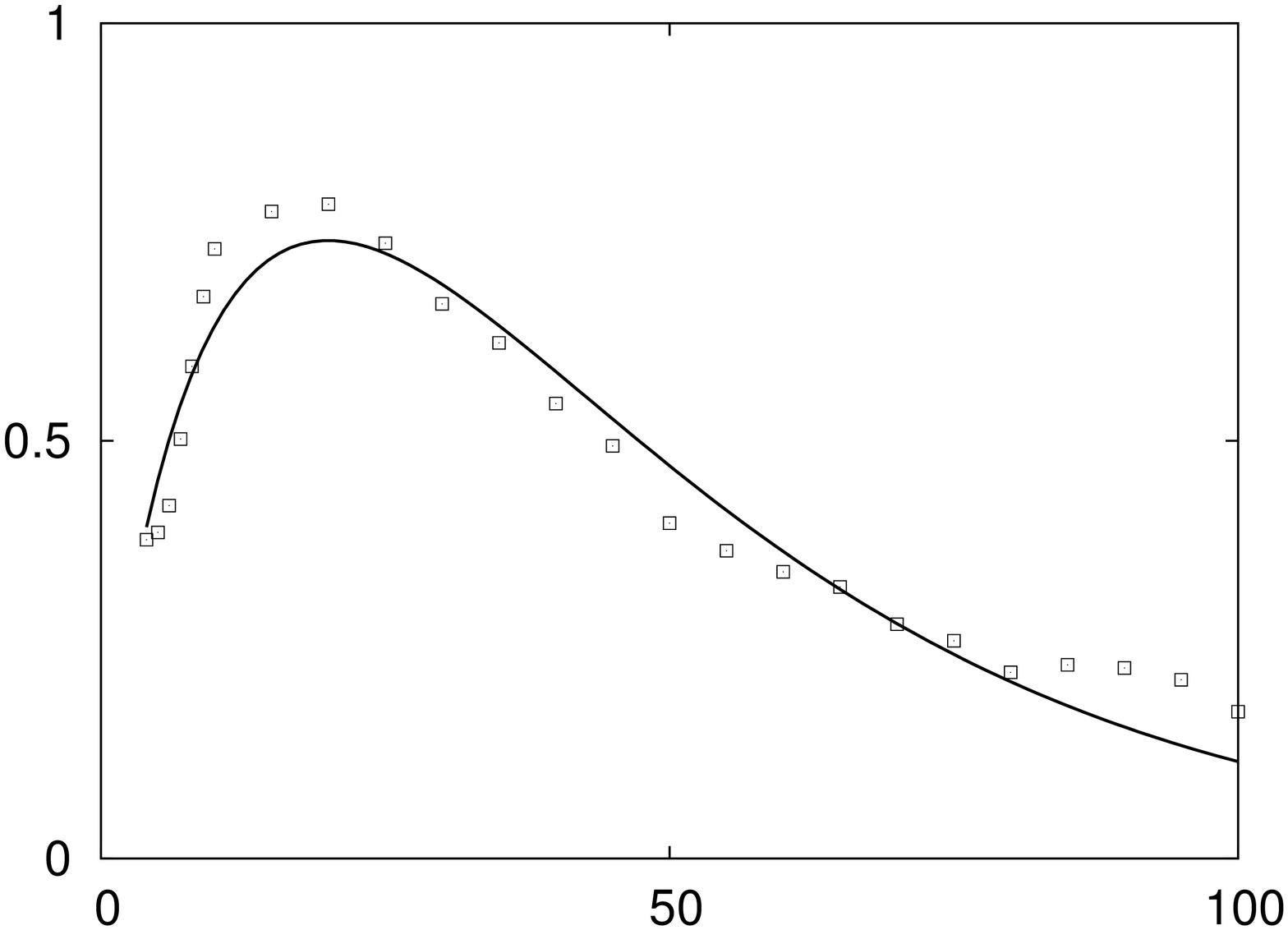}}
      \put(-3,46){$\Pi$}
      \put(58,0){$N$}
      %@generator gnuplot
      %@script figs/data/performance/symBreak/symBreakScan.plt
      %@data figs/data/performance/symBreak/symBreakScan*
      %\graphpaper[10](0,0)(120,90)
    \end{picture}
  }
  \hspace{1.5mm}
  \subfigure[\label{fig:fitted:diplomarbeit}Aggregation in tree-like structures and reduction to shortest path~\citep{hamann06}; $\Pi$~gives the ratio of successful runs.]{
    \begin{picture}(120,85)
    \put(-3,88){\includegraphics[angle=270,width=120\unitlength]
        {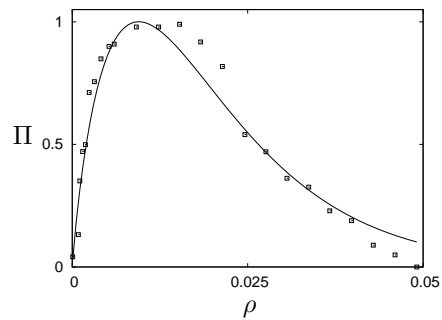}}
      \put(-3,46){$\Pi$}
      \put(60,0){$\rho$}
      %@generator gnuplot
      %@script figs/data/performance/diplomarbeit/diplomarbeit.plt
      %@data figs/data/performance/diplomarbeit/diplomarbeitData
      %\graphpaper[10](0,0)(120,90)
    \end{picture}
  }
  \caption{\label{fig:fitted}Model of cooperation and interference and
    three scenarios with fitted performance $\Pi$ according to
    Eq.~\ref{eq:perf:fullEq}}
\end{figure}

We analyze the forth example, emergent
taxis~\citep{nembrini02,bjerknes07}, in more detail and, for this purpose,
give a short description of the algorithm. The objective is to move a
swarm of robots towards a light beacon. The robots are limited in
their capabilities because they only have an omnidirectional beacon
sensor that detects two states, beacon seen or not seen, but
gives no bearing. If a robot does not see the beacon this is always
because one or several other robots are within the line of sight
between this robot and the beacon. Robots that see the beacon are
called `illuminated' and robots that do not see the beacon are called
`shadowed'. The robots' behavior is defined to differ depending on
these two states. Shadowed robots have a shorter collision avoidance
radius than illuminated robots. Consequently if a shadowed and an
illuminated robot approach each other, the illuminated robot will
trigger its collision avoidance behavior while the shadowed robot will
not be affected until it gets even closer and triggers its own
collision avoidance behavior as well. This interplay between shadowed
and illuminated robots generates a bias towards the beacon. In
addition, the algorithm has a `coherence state' which aims at keeping
the robots together within each others communication range. It is
assumed that a robot is able to count the robots that are within
range. Once this number drops below a threshold~$\alpha$, the robot
will do a u-turn which hopefully brings it back into the swarm. In the
following investigations we set the threshold at first to~$\alpha=0$
which means we turn the coherence behavior off. Later we
follow~\citep{bjerknes07} and set the threshold to swarm size
($\alpha=N$) to enforce full coherence. Initially the robot swarm is
distant from the beacon and is randomly distributed while ensuring
coherence. Interestingly, it is difficult to identify two separated
behavior components of cooperation and interference. The robots
cooperate in generating coherence and in having shadowed robots that
approach illuminated robots which drives them towards the
beacon. However, collision avoidance behavior also has disadvantageous
effects. In order to keep coherence the robots might be forced to
aggregate too densely which might result in robots blocking each
other.

The following empirical data is based on a simple simulation of
emergent taxis. This simulation is noise-free and therefore robots
move in straight lines except for u-turns according to the emergent
taxis algorithm (a~random turn after regaining coherence was not
implemented).

First, we measure the performance that is achieved without
cooperation. This is done by defining a random behavior that ignores
any characteristic feature of the actual emergent taxis algorithm. We
set this parameter to $\alpha=0$ in the simulation to obtain the
cooperation-free behavior. Hence, no robot will ever u-turn and they
basically disperse in the arena. A~simulation run is stopped once a
robot touches a wall. The performance~$\Pi$ of the swarm is measured by
the total distance covered by the swarm's barycenter multiplied by the
swarm size (i.e., an estimate of the sum of all distances that were
effectively covered by each robot). The performance obtained by this
random behavior can be fitted using the interference function of
Eq.~\ref{eq:perf:interference} (interpreting the interference function
as a measurement of swarm performance without cooperation). The fitted
interference function and the empirically obtained data is shown in
Fig.~\ref{fig:taxis:fitted} labeled `random'. The interference
function does not drop to zero for large~$N$; this bias towards the
light source (positive covered distance) is due to the initial
positioning of the swarm closer to the wall that is farther away from
the light source. 

In a second step, the full model of swarm performance~$\Pi$
(Eq.~\ref{eq:perf:fullEq}) is to be fitted to the actual emergent
taxis scenario. The data is obtained by setting the threshold in our
simple simulation of emergent taxis back to normal ($\alpha=N$). The
fitting is done by keeping the interference function fixed and we fit
only the cooperation function (i.e., fitting $a_1$ and $b$ while
keeping $a_2$, $c$ and $d$ fixed). The fitted swarm performance model
is shown in Fig.~\ref{fig:taxis:fitted} labeled `emergent taxis'. This
simple model is capable of predictions, if the interference function
has been fitted and we fit the cooperation function only to a small
interval of, for example, $N\in [20,22]$ (i.e., intervals close to the
maximal performance). This is shown in
Fig.~\ref{fig:taxis:narrowFit}. The implication is that if the
interference function is known as well as the optimal swarm size then
the behavior within the other intervals can be predicted.

\setlength{\unitlength}{0.0039\textwidth}
\begin{figure}
  \centering
  \subfigure[\label{fig:taxis:fitted}Model fitted to data of random and emergent--taxis behavior; performance~$\Pi$ is the total distance covered by the swarm's barycenter multiplied by the swarm size~$N$.]{
    \begin{picture}(110,85)
    \put(-4,88){\includegraphics[angle=270,width=120\unitlength]
        {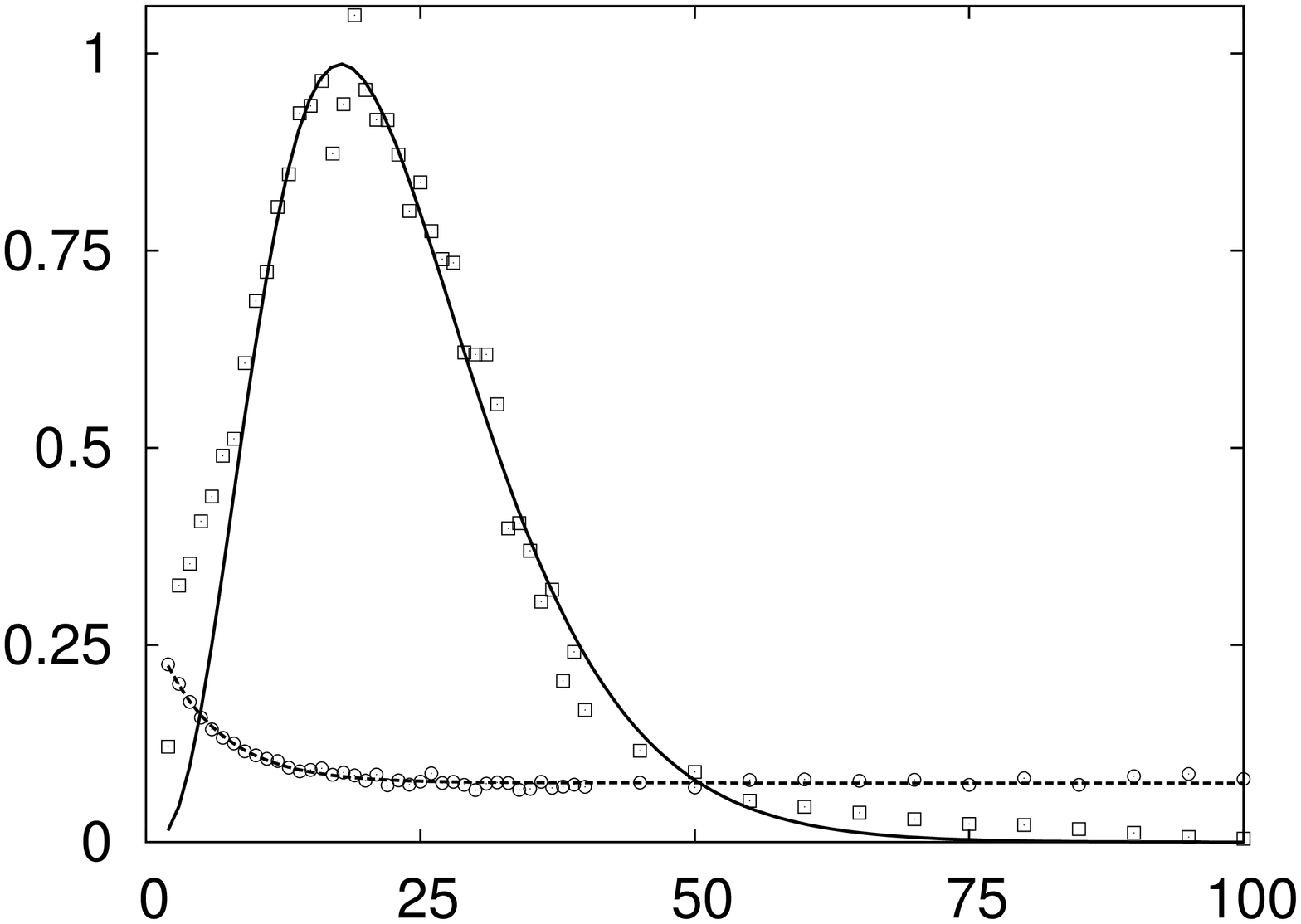}}
      \put(-1,44){$\Pi$}
      \put(60,0){$N$}
      \put(80,20){random}
      \put(46,60){emergent taxis}
      %@generator gnuplot
      %@script figs/data/performance/taxis/taxisSuccessRate.plt
      %@data figs/data/performance/taxis/taxisSuccessRate*
      %\graphpaper[10](0,0)(110,85)
    \end{picture}
  }
  \hspace{1.5mm}
  \subfigure[\label{fig:taxis:narrowFit}Fit of the cooperation function to the emergent--taxis performance based on the narrow interval of $N\in{[20,22]}$ only, with fixed parameters of interference function as shown in (a).]{
    \begin{picture}(110,85)
    \put(-4,88){\includegraphics[angle=270,width=120\unitlength]
        {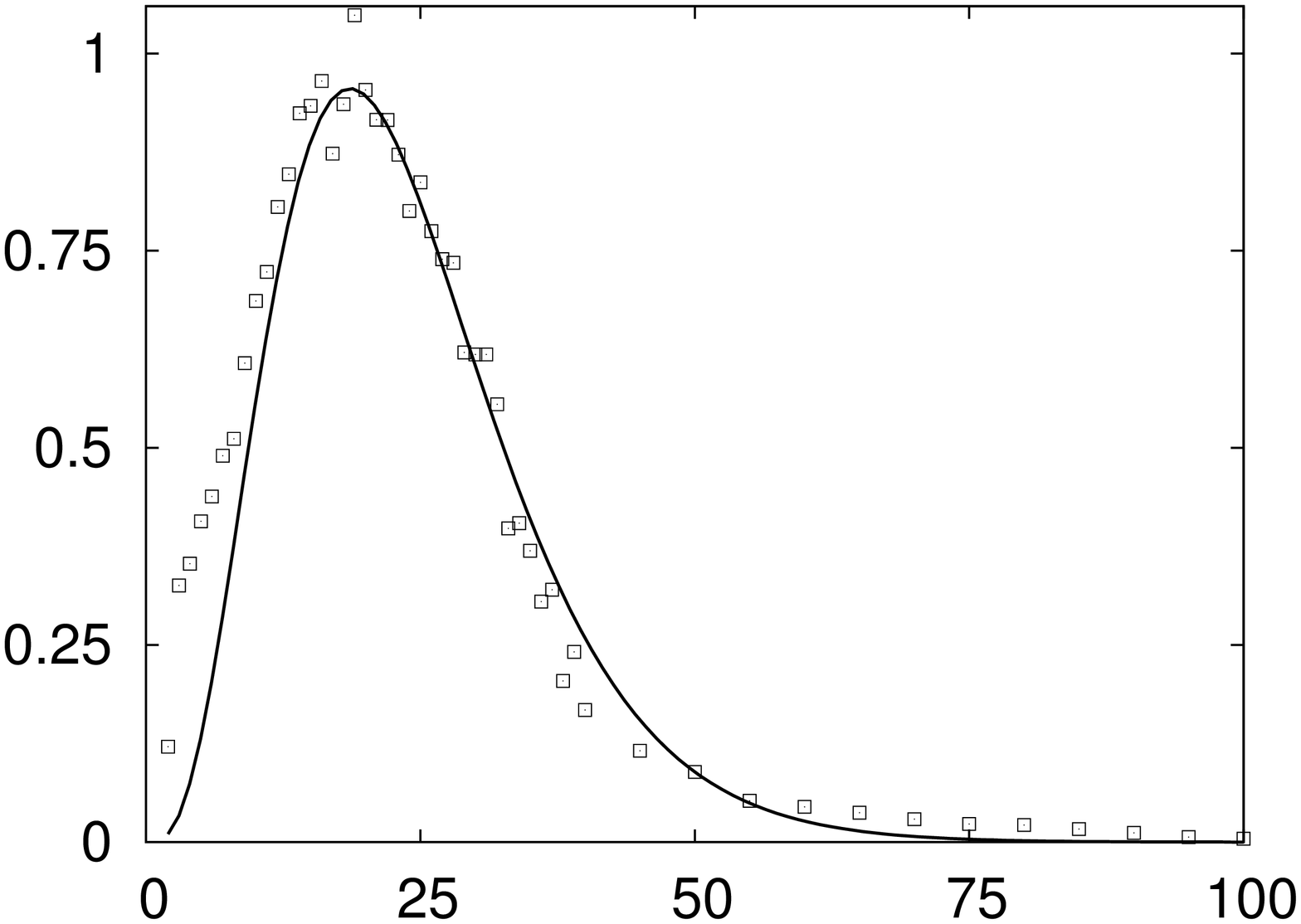}}
      \put(-1,44){$\Pi$}
      \put(60,0){$N$}
      %\put(80,20){random}
      %\put(46,60){emergent taxis}
      %@generator gnuplot
      %@script figs/data/performance/taxis/taxisSuccessRate.plt
      %@data figs/data/performance/taxis/taxisSuccessRate*
      %\graphpaper[10](0,0)(110,85)
    \end{picture}
  }
  \subfigure[\label{fig:taxisHist2d}Histogram of barycenter speeds~$v$
    for different swarm sizes. Note the bimodality in the interval
    $N\in {[}16,40{]}$.]{
    \begin{picture}(125,70)
    \put(0,85){\includegraphics[angle=270,width=130\unitlength]
      {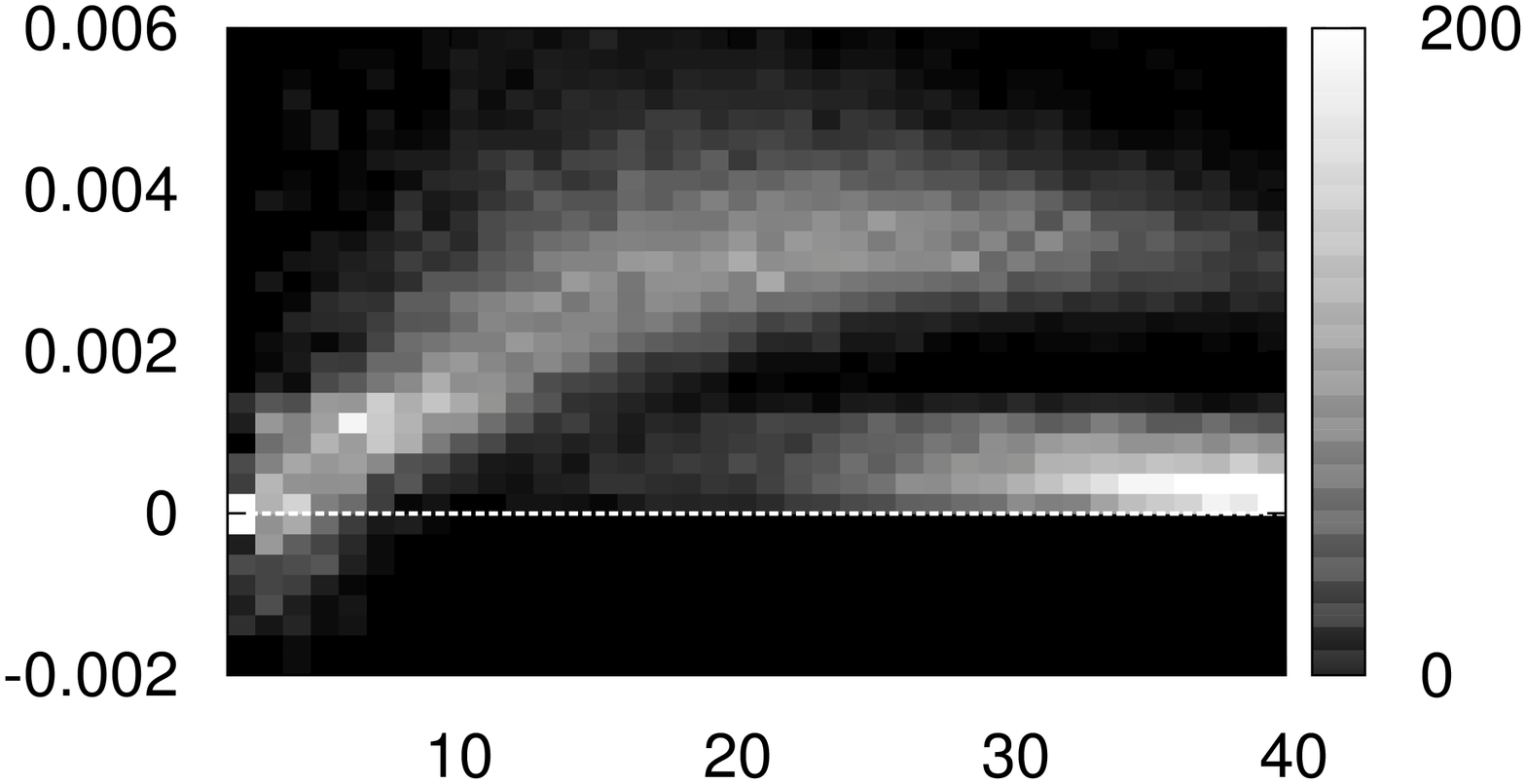}}
    \put(60,0){\small $N$}
    \put(-1,42){\small $v$}
    \put(121,27){\small \begin{sideways}frequency\end{sideways}}
    % @generator gnuplot
    % @script figs/data/performance/taxis/hist2D.plt
    % @data figs/data/performance/taxis/hist2dData
    %\graphpaper[10](0,0)(125,70)
  \end{picture}
  }
  \caption{\label{fig:taxis}Performance of a random behavior and the
    actual self-organized emergent taxis behavior (also sometimes
    called `alpha algorithm')~\citep{nembrini02,bjerknes07} with fitted
    model (Eq.~\ref{eq:perf:fullEq}); histogram showing two phases.}
\end{figure}

Note that the model operates on a single averaged value to describe
the performance which does not fully catch the system's behavior. At
least in some scenarios, as here in the emergent taxis scenario, the
performance does not just continuously decrease due to continuously
increasing interference. Instead, two coexisting phases of behaviors
emerge: functioning swarms moving forward and pinned swarms with
extreme numbers of u-turns. In emergent taxis this is shown, for
example, by a histogram of barycenter speeds in
Fig.~\ref{fig:taxisHist2d}. For $N<15$ the mean of a unimodal
distribution increases with increasing~$N$. Starting at about $N=15$ a
second phase emerges that shows slowly moving swarms and generates a
bimodal distribution. Hence, given the fully deterministic
implementation of our simulation, there are two classes of initial
states (robot positions and orientations) that determine the two
extremes of success or total failure. Consequently, swarm performance
functions as shown in Fig.~\ref{fig:taxis:fitted} need to be
interpreted with care because they might indicate an average behavior
that does actually not occur. Still, these swarm performance functions
are useful if the values are interpreted relatively as success
rates. That way Fig.~\ref{fig:taxis:fitted} gives a good estimate of
the frequency of the two phases. In other scenarios the interference
might truly increase continuously due to a qualitatively different
process, such as saturation of target areas with robots.

The presented model of swarm performance has potential to be
applicable to many swarm systems. In the next section, a model is
given for a subset of swarm systems namely collective decision-making
systems. The two models relate to each other as in some cases they are
both applicable to the same swarm system. A~candidate for such a
system could be a BEECLUST-controlled swarm. The application of the
swarm performance model to this system is given in
Fig.~\ref{fig:fitted:symBreakScan}. Data that supports, that an
application of the following collective decision model is likely, is
given by \citet{hamann10a}. In such systems the effectivity of the
collective decision and the performance of the swarm are directly
linked and consequently the two reported models are, too.

\section{Universal properties of collective decisions}
\label{sec:collectiveDecisions} 

In the following, we investigate macroscopic models of collective
decisions. One of the most general and at the same time simplest
models of collective decisions is a model of only one state
variable~$s(t)$, which gives the temporal evolution of the swarm
fraction that is in favor of one of the options in a binary decision
process. If we assume that there is no initial bias to either option
(i.e., full symmetry), then we need a tie breaker for $s=0.5$. A~good
choice for a tie breaker is noise because any real swarm will be
noisy. The average change of~$s$ depending on itself per time~($\Delta
s(s)/\Delta t$) is of interest. Given that the system should be able
to converge to either of the options at a time plus having the
symmetric case of~$s=0.5$, function~$\Delta s(s)/\Delta t$ needs to
have at least three roots ($\Delta s(s_i)/\Delta t=0$, $s_i\in S$,
$|S|\ge 3$) and consequently is at least a cubic function. Instead of
developing a model, that defines such a function, we prefer a model
that allows this function to emerge from a simple process. Once
symmetry is overcome by fluctuations, swarm systems have a tendency to
confirm and reinforce such a preliminary decision due to positive
feedback (say, for $s=0.5+\epsilon$ there is a tendency towards
$s=1$). We define such a process depending on probabilities of
positive feedback next.

\subsection{Simple model of collective decisions}
\label{sec:modelOfCollDec}

As discussed in the introduction, we define an urn model that was
inspired by the models of \citet{ehrenfest07} and of
\citet{eigen93}. We use an urn model that has optionally positive or
negative feedback depending on the system's current state and
depending on a stochastic process. The urn is filled with $N$~marbles
which are either red or blue. The game's dynamics is
turn-based. First, a marble is drawn with replacement followed by
replacing a second one determined by the color of the first
marble. The probability of drawing a blue marble is implicitly
determined by the current number of blue marbles in the urn. The
subsequent replacement of a second marble has either a positive or a
negative feedback. Say, we draw a blue marble, we notice the color,
and put it back into the urn. Then our model defines that with
probability~$P$ a red marble will be replaced by a blue one (i.e.,
a~positive feedback event because drawing a blue one increased the
number of blue marbles) and with probability~$1-P$ a blue one will be
replaced by a red one (i.e., a~negative feedback event because now
drawing a blue one decreased the number of blue marbles). Hence,
$P$~gives the probability of positive feedback.

The analogy of this model to a collective decision making scenario is
the following. The initial drawing resembles the frequency of
individual decisions in the swarm over time within the turn-based
model. This frequency is proportional to the current ratio~$s(t)$ of
blue marbles in the urn. Consequently, we serialize the system
dynamics and each system state~$s(t)$ has at most two predecessors
($s(t-1)=s(t)\pm 1$). The replacement of the second marble resembles
the effect of either a swarm member convincing another one about its
decision or being convinced of the opposite.

The probability of positive feedback~$P$ is determined explicitly. We
define the determination whether positive feedback or negative
feedback is effective in a given system state~$s$ as a binary random
experiment. The sample space is $\Omega = \{\text{PFB},\text{NFB}\}$
with PFB denoting positive feedback, NFB denoting negative feedback,
and we define a probability measure~$m$, consequently
$m(\text{PFB})+m(\text{NFB})=1$ holds. Hence, the probability of
positive feedback is defined by~$m(\text{PFB})=P(s,\varphi)$
(probability of negative feedback is $m(\text{NFB})=1-P(s,\varphi)$)
and for now we choose a sine function
\begin{equation}
  %  P(s) = \frac{1}{2} - \frac{1}{2}\cos\left(c\pi|\frac{1}{2} - s|\right)
  \label{eq:posFBsine}
  P(s,\varphi) = \varphi\sin(\pi s) \enspace .
\end{equation}
Due to the symmetry $P(s,\varphi)=P(1-s,\varphi)$ it is irrelevant
whether $s$ is set to the ratio of blue marbles or the ratio of red
marbles. Later we will find that similar but different functions might
be a better choice in certain
situations~(Sec.~\ref{collectiveDecisions:examples}). The
constant~$\varphi\in[0,1]$ scales the amplitude of the sine function
(see Fig.~\ref{fig:symBreak:probs}) and consequently defines the
predominant `sign' of the feedback and the probabilities of positive
and negative feedback. The integral $\int_0^1P(s,\varphi)ds$ gives the
overall probability of having positive feedback independent
of~$s$. Negative feedback is predominant for any~$s$ for $\varphi <
0.5$ and an interval around~$s=0.5$ emerges for which positive
feedback is predominant for $\varphi > 0.5$.

\setlength{\unitlength}{0.00371\textwidth}
\begin{figure}%[h]
  \centering
  \subfigure[\label{fig:symBreak:probs}Examples of setting the probability of positive feedback for intensities of feedback~$\varphi\in\{0,0.125,0.25,0.5,0.75\}$, squares give the redetermined values according to Eq.~\ref{eq:redeterminFB}.]{
    \begin{picture}(120,90)
      \put(0,95){\includegraphics[angle=270,width=130\unitlength]
        {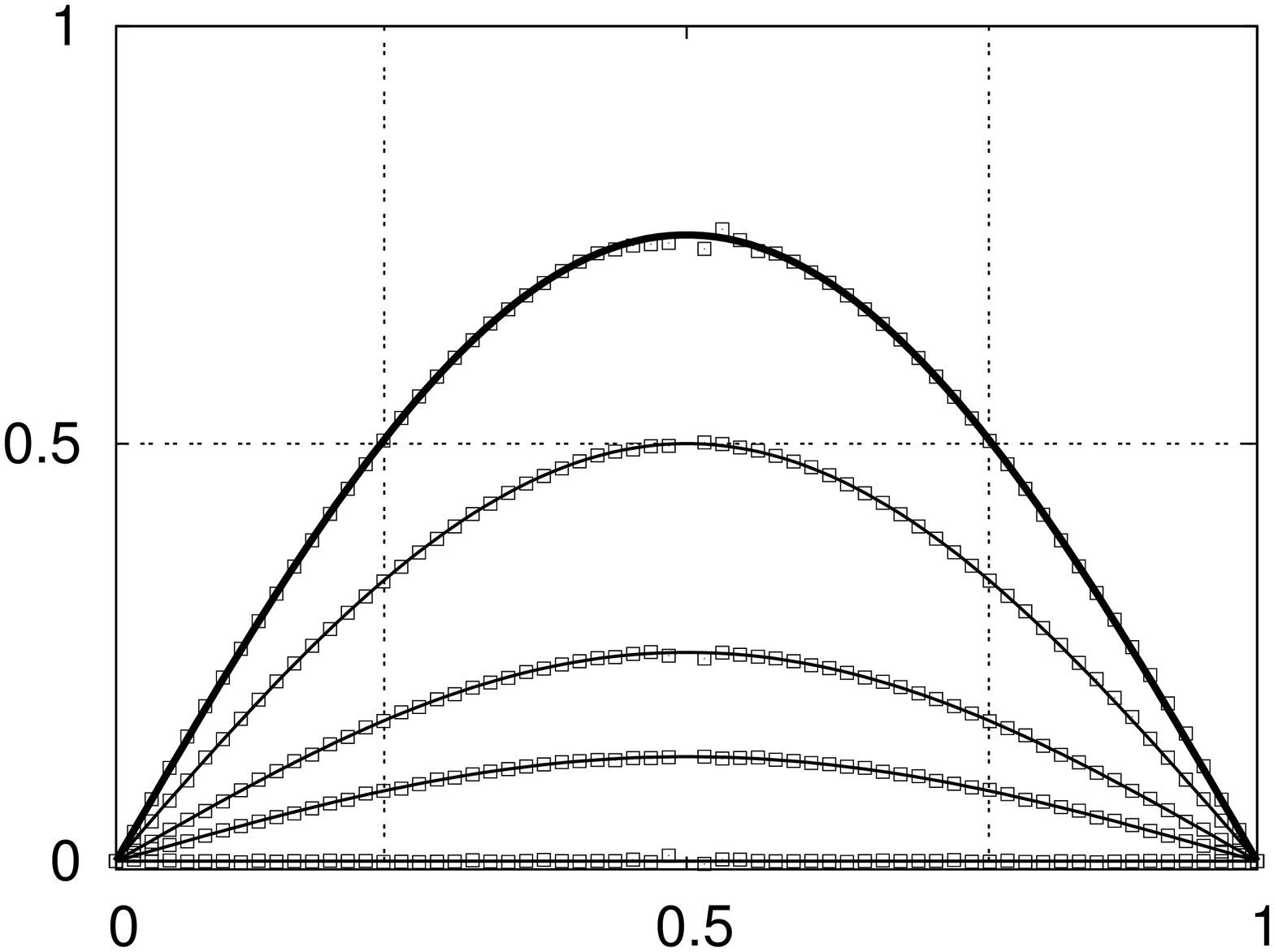}}
      \put(69,1){\small $s$}
      \put(41.5,10){\scriptsize $s_1$}
      \put(95,10){\scriptsize $s_2$}
      \put(0,50){\small $P$}
      \put(61,17){\scriptsize $\varphi=0$}
      \put(55,26){\scriptsize $\varphi=0.125$}
      \put(57,35){\scriptsize $\varphi=0.25$}
      \put(59,54){\scriptsize $\varphi=0.5$}
      \put(57,73){\scriptsize $\varphi=0.75$}
      % @generator gnuplot
      % @script figs/data/urnModel/symBreakChange.plt
      % \graphpaper[10](0,0)(120,90)
    \end{picture}
  }
  \hspace{1.5mm}
  \subfigure[\label{fig:symBreak:change}Average change of $B(t)$
in terms of marbles, lines according to Eq.~\ref{eq:DeltaB}, squares give empirical data, number of samples is~$8\times 10^5$ for each possible~$s$, 64 marbles.]{
    \begin{picture}(130,90)
      \put(0,95){\includegraphics[angle=270,width=130\unitlength]
        {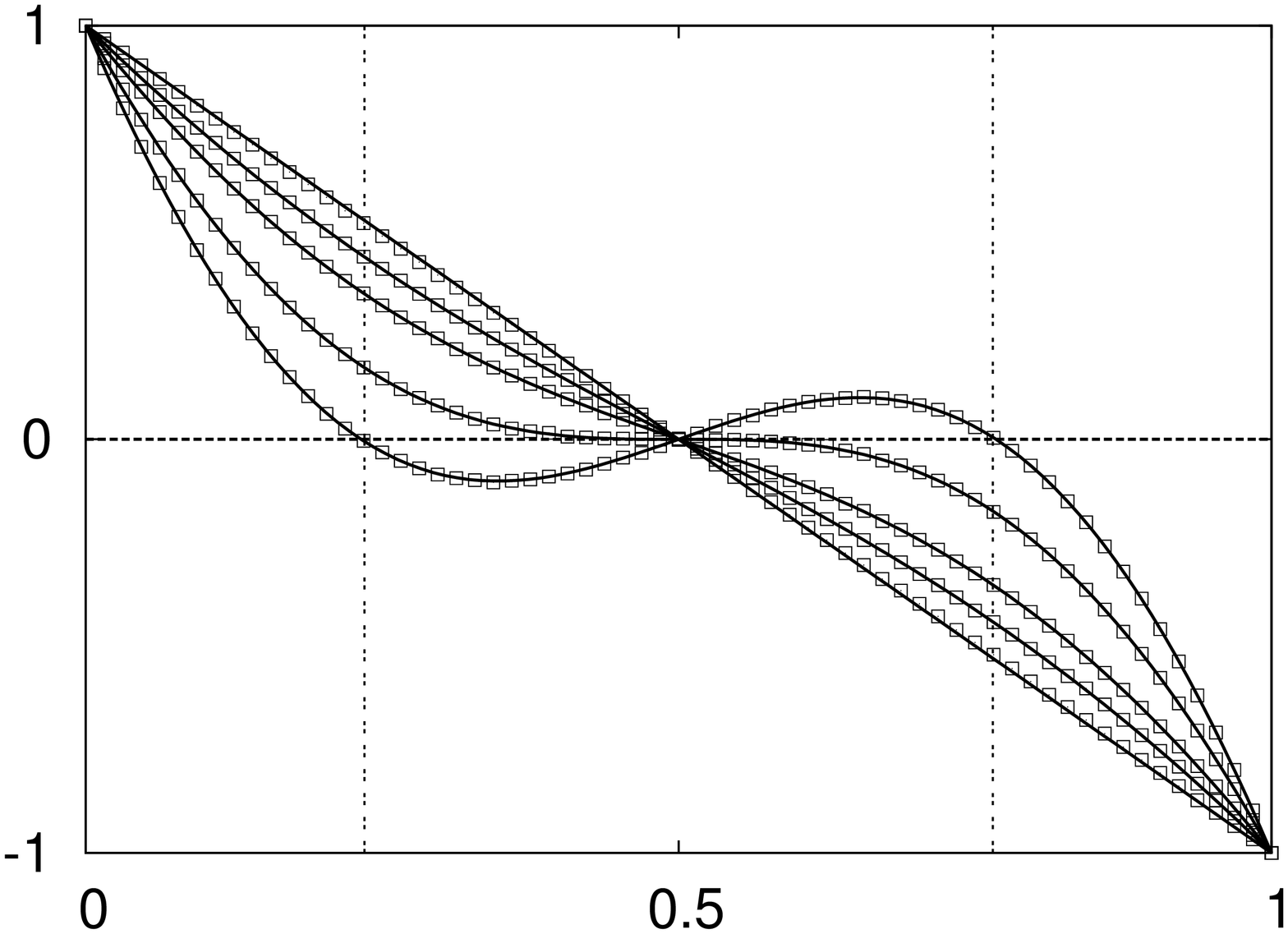}}
      \put(66,1){\small $s$}
      \put(36,10){\scriptsize $s_1$}
      \put(93,10){\scriptsize $s_2$}
      \put(-2,32){\begin{sideways}\small $\Delta B$ and $\overline{\Delta B}$\end{sideways}} %\overline{\frac{B(t)}{\Delta t}}$}
      \put(71,32){\scriptsize $\varphi=0$}
      \put(97,20){\line(1,2){5}}
      \put(81,16){\scriptsize $\varphi=0.125$}
      \put(85,44){\line(0,1){20}}
      \put(76,67){\scriptsize $\varphi=0.25$}
      \put(100,41){\line(1,2){7.5}}
      \put(99,57){\scriptsize $\varphi=0.5$}
      \put(25,40){\scriptsize $\varphi=0.75$}
      % @generator gnuplot
      % @script figs/data/urnModel/symBreakChange.plt
      % @data figs/data/urnModel/hamannChange_sin_*
      %\graphpaper[10](0,0)(120,90)
    \end{picture}
  }
  \caption{\label{fig:symBreakFirstThird}Settings of the positive feedback
    probabilities and resulting average change in $B(t)$ in the urn model
    over the ratio of marbles~$s$.}
\end{figure}

$P(s,\varphi)$~is plotted for different settings of~$\varphi$ in
Fig.~\ref{fig:symBreak:probs}. There is maximum probability for
positive feedback for the fully symmetric case of $s=0.5$ as clearly
seen in Fig.~\ref{fig:symBreak:probs}. For $s=0$ and $s=1$ we have
$P(s,\varphi)=0$ because no positive feedback is possible (either all
marbles are already blue or all marbles are red and therefore no blue
one can be drawn). For~$\varphi\le 0.5$ the probability of positive
feedback is small ($\varphi\le 0.5,\forall s:P(s,\varphi)\le 0.5$),
consequently the system is stable and kept around $s=0.5$.

We can calculate now the average expected change per round~$\Delta B$
of blue marbles~$B$ by summing over the four cases: drawing a blue or
red marble, followed by positive or negative feedback, multiplied by
the `payoff' in terms of blue marbles, respectively. Using the
symmetry~$P(s,\varphi)=P(1-s,\varphi)$ we get
\begin{align}
  \Delta B(s) &=&& sP(s,\varphi)(+1)+s(1-P(s,\varphi))(-1)\notag\\
  &&&+(1-s)P(1-s,\varphi)(-1)+(1-s)(1-P(1-s,\varphi))(+1)\notag\\
  &=&& 4(P(s,\varphi)-0.5)(s-0.5)\label{eq:DeltaB} \enspace .
\end{align}

We defined $P(s,\varphi)$ based on a trigonometric function but
alternatively one can choose, for example, a quadratic function
\begin{equation}
\label{eq:posFBquadr}
P(s,\varphi)=\varphi(1-4(s-0.5)^2) \enspace ,
\end{equation}
yielding a cubic function 
\begin{equation}
\Delta B(s)=-2(s-0.5)+4\varphi(s-0.5)-16\varphi(s-0.5)^3 \enspace .
\end{equation}
In the following, however, we will use Eq.~\ref{eq:DeltaB}. Also note
that by turning positive feedback completely off ($P(s,\varphi)=0$) we
obtain the Ehrenfest urn model again
(cf. Eqs.~\ref{eq:ehrenfest:deltaB} and \ref{eq:DeltaB}) and by
activating maximal positive feedback ($P(s,\varphi)=1$) we obtain the
Eigen urn model (cf. Eq.~\ref{eq:eigen:deltaB}).

While in the above definition the positive feedback
probability~$P(s,\varphi)$ is explicitly given, it might be unknown in
applications of this model. In these applications, the positive
feedback itself might not be measurable but maybe the number of
observed decision revisions of the agents. We introduce the absolute
number of observed individual decision revisions from red to
blue~$r_b(s)$ over any given period and from blue to red~$r_r(s)$. The
ratio of red-to-blue revisions is directly related to the expected
average change per round. Assuming payoffs of~$\pm 1$, the average
change of blue marbles per round is obtained by the weighted sum
\begin{equation}
\frac{r_b(s)}{r_b(s)+r_r(s)}\cdot (+1) + \frac{r_r(s)}{r_b(s)+r_r(s)}\cdot (-1) =\Delta B(s) \enspace ,
\end{equation}
which simplifies to
\begin{equation}
\label{eq:switchingRatio}
\frac{r_b(s)}{r_b(s)+r_r(s)}=0.5(\Delta B(s)+1) \enspace .
\end{equation}
In addition, the ratio~$\frac{r_b(s)}{r_b(s)+r_r(s)}$ also directly
relates to the positive feedback probability~$P(s)$ by considering the
above mentioned four cases: drawing a blue or red marble, followed by
positive or negative feedback. Red-to-blue revisions are observed only
in two of these four cases: drawing a blue marble followed by positive
feedback and drawing a red marble followed by negative feedback. The
summed probability of these two cases has to equal the ratio of
red-to-blue revisions, which is consequently interpreted as
probability. The equation
\begin{equation}
\label{eq:redeterminFBAnsatz}
P(s)s+(1-P(1-s))(1-s)=\frac{r_b(s)}{r_b(s)+r_r(s)} \enspace ,
\end{equation}
yields using the symmetry $P(1-s)=P(s)$
\begin{equation}
\label{eq:redeterminFB}
P(s)=\frac{\frac{r_b(s)}{r_b(s)+r_r(s)}-1+s}{2s-1}, \text{ for }s\ne 0.5, \frac{r_b(s)}{r_b(s)+r_r(s)}\le \max(s,1-s) \enspace ,
\end{equation}
an equation which is to be used to determine the positive feedback
probability based on measurements of agents' revisions. The pole
at~$s=0.5$ and the consequently undefined~$P(s=0.5)$ is
reasonable. Any definition of $P(s=0.5)$ would be without effect to
the system because negative and positive feedback are
indistinguishable at $s=0.5$. We actually define $P(s=0.5)$ in
Eq.~\ref{eq:posFBsine}, which is, however, only a simplification of
notation. Also note that this mathematical difficulty has limited
effect in applications because swarms are inherently discrete. For
example, $s=0.5$ does not exist for odd swarm sizes. The squares in
Fig.~\ref{fig:symBreak:probs} give the redetermined~$P(s)$ based on
Eq.~\ref{eq:redeterminFB} and on measurements of $r_b(s)$ and
$r_r(s)$. The values for~$P(s=0.5)$ cannot be given, as discussed, and
also values in the vicinity of~$s=0.5$ show discrepancies because they
are close to the pole of Eq.~\ref{eq:redeterminFB}.

In Fig.~\ref{fig:symBreak:change} we compare the theoretical average
change per round~$\Delta B$ according to Eq.~\ref{eq:DeltaB} to the
empirically obtained average change~$\overline{\Delta B}$ in terms of
number of marbles for the different settings of~$\varphi$. The
agreement between theory and empirical data is good (root mean
squared errors of~$<6\times 10^{-3}$) as expected. Two zeros~$s_1$ and
$s_2$ emerge additionally to $s_0=0.5$ for $\varphi>0.5$:
$s_1=\frac{1}{\pi}\arcsin(\frac{1}{2\varphi})$ and
$s_2=1-\frac{1}{\pi}\arcsin(\frac{1}{2\varphi})$. Positive values
of~$\Delta B(s)$ for $s<0.5$ represent dynamics that has a bias
towards $s=0.5$ and negative values represent dynamics with a bias
towards $s=0$ and vice versa for the other half ($s>0.5$).

\setlength{\unitlength}{0.004\textwidth}
\begin{figure}%[h]
  \centering
  \begin{picture}(130,90)
    \put(-7,100){\includegraphics[angle=270,width=140\unitlength]
      {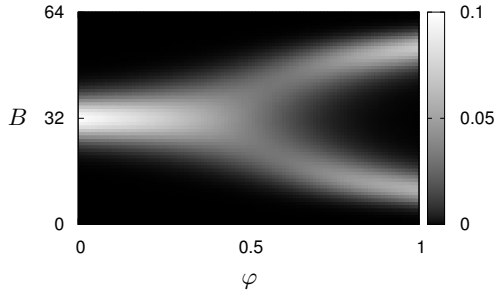}}
    \put(61,8){\small $\varphi$}
    \put(-3,51.5){\small $B$}
    % @generator gnuplot
    % @script figs/data/urnModel/histo.plt
    % @data figs/data/urnModel/histo
    % \graphpaper[10](0,0)(130,90)
  \end{picture}
  \caption{\label{fig:symBreak:histo}Normalized histogram of blue
    marbles~$B$ over intensity of feedback~$\varphi$ after $t=2000$
    steps in the urn model, initialized to $B(0)\in\{32,33\}$,
    indicating a pitchfork bifurcation at ~$\varphi=0.5$.}
\end{figure}

Fig.~\ref{fig:symBreak:histo} gives an estimate of the asymptotic
behavior of this urn model for varied feedback intensity~$\varphi$. It
shows a pitchfork bifurcation at~$\varphi=0.5$, which is to be
expected based on Fig.~\ref{fig:symBreak:change} and which is fuzzy
because of the underlying stochastic process. For $\varphi>0.5$, the
curve defined by~$\Delta B(s)$ becomes cubic and generates two new
stable fixed points while the former at~$s=0.5$ becomes unstable.

\subsection{Optional extension of the model}

Since positive and negative feedback is determined stochastically
based on the current global state in the urn model, it is
straightforward to determine the number of replaced marbles
stochastically, too. Instead of having a fixed `payoff' (number of
replaced marbles) of $+1$ for positive feedback and $-1$ for negative
feedback we can, for example, define a probability~$M$ of the event
`having a payoff of~$+1$' (or $-1$ respectively) and a probability
$1-M$ for a payoff of~0. Thus the average payoff would be~$+M$ and
$-M$ for positive and negative feedback respectively. In addition, the
payoff can be variant depending on the current global state~$s$, that
is, we define a function~$M(s)$. It turns out that a definition
similar to the positive feedback probability~$P(s,\varphi)$ is useful
here. We define the variant payoff by
\begin{equation}
  \label{eq:payoff}
  M(s) = c_1\sin(\pi s)+c_2 \enspace ,
\end{equation}
for appropriately chosen constants $c_1$ and $c_2$. $M(s)$ defines the
average over absolute values of changes in~$s$, similar to the
diffusion coefficient in Fokker-Planck theory. Measurements of the
diffusion coefficient in swarm systems were reported
in~\citet{yates09} and \citet{hamann10c}, which show low values for
$s\ll 0.5$ and $s\gg 0.5$ with a peak at $s=0.5$. With $M(s)$ defined
by Eq.~\ref{eq:payoff} we have symmetry again ($M(s)=M(1-s)$) and
hence the extension of Eq.~\ref{eq:DeltaB} is simply
\begin{equation}
  \label{eq:DeltaBextended}
  \Delta B(s)=4M(s)(P(s,\varphi)-0.5)(s-0.5) \enspace .
\end{equation}

\subsection{Available and unavailable methods}
\label{sec:methods} 

Note that the recurrence $B_t=B_{t-1}+\Delta B(B_{t-1})$ is a
trigonometric or cubic function based on Eq.~\ref{eq:posFBsine} or
Eq.~\ref{eq:posFBquadr}, respectively. Thus, it is much more difficult
to be handled analytically and a concise result as for the Ehrenfest
model (Eq.~\ref{eq:ehrenfestEvolutionAnalytically}) cannot be
obtained. When applying nonlinear equations for $\Delta B$ we enter
the domains of nonlinear dynamics and chaotic systems with all their
known mathematical intractabilities. Hence, in general we have to rely
on numerical methods.

However, if we choose to investigate probability
distributions~$\rho(s,t)$ instead of single trajectories~$s(t)$, an
interesting mathematical option is available. The steady
state~$\pi(s)$ of the probability distribution~$\rho(s,t)$ over an
ensemble of realizations of~$s(t)$ can be obtained analytically. We
assume for simplicity that the current state of the collective
decision system changes every step by exactly one marble. The
process defined by the urn model is memoryless, that is, it has the
Markov property and we can define a Markov chain with $N+1$ states. We
define the transition matrix {\bf T} of the Markov chain by
\begin{align}
  \label{eq:transitionMatrix}
  T_{2,1}&=\frac{1}{2}\left(\Delta B\left(\frac{0}{N}\right)+1\right),&\notag\\
  T_{i+1,i}&=\frac{1}{2}\left(\Delta B\left(\frac{i-1}{N}\right)+1\right),&\text{ for }i\in[2,N],\notag\\
  T_{i-1,i}&=1-\frac{1}{2}\left(\Delta B\left(\frac{i-1}{N}\right)+1\right),&\text{ for }i\in[2,N],\notag\\
  T_{N,N+1}&=1-\frac{1}{2}\left(\Delta B\left(\frac{N}{N}\right)+1\right) \enspace .&
\end{align}
The steady state is then computed by determining the eigenvectors ${\bf v}$
\begin{equation}
  \label{eq:eigenvectors}
  {\bf Tv}=\lambda {\bf v} \enspace .
\end{equation}
Generally there are several eigenvectors but only one that has no
changing sign in its elements (all positive or all negative) which
represents the equilibrium distribution of the Markov chain~$\pi$ or
$-\pi$ respectively. 

With the steady state at hand, several methods of statistical analysis
are available. As indicated by the data obtained by simulation shown
in Fig.~\ref{fig:symBreak:histo}, the equilibrium distributions are
bimodal for $\varphi>0.5$. Hence, it is an option to calculate
splitting probabilities~\citep{gardiner85}. The splitting probability
$\sigma_{a,b}(x)$ gives the probability that the system initialized at
$s=x$ will reach the state $s=b$ before the state $s=a$. It is
calculated by
\begin{equation}
  \sigma_{a,b}(x)=\int_a^x \pi(s)^{-1}ds \left( \int_a^b \pi(s)^{-1}ds \right)^{-1} \enspace .
\end{equation}
Note that this equation is based on a continuous distribution~$\pi(s)$
which can be obtained from a discrete distribution (e.g., the above
equilibrium distribution based on Markov theory) by fitting. Another
option, which was used here, is to apply Fokker--Planck theory which
allows to calculate a continuous equilibrium distribution directly
based on $\Delta B(s)$~\citep{hamann10c}. We set $a$ and $b$ to the
positions of the two peaks in the steady state probability
density~$\pi$ that we obtain for $\varphi=1$. Results for varied
positive feedback intensity~$\varphi$ (while keeping $a$ and $b$
constant) are shown in Fig.~\ref{fig:split}. It is clearly seen that
for $\varphi<0.5$ there is a wide interval with a fifty-fifty chance
to end at either~$a$ or~$b$. This is because the steady state
probability density~$\pi$ is unimodal for~$\varphi<0.5$. Beginning
with $\varphi=0.5$, when the steady state probability density starts
to be bimodal, and with increasing $\varphi$ the probability of
switching from one peak to the other is decreasing considerably which
is an indicator for an effective collective decision.

\setlength{\unitlength}{0.0045\textwidth}
\begin{figure}
  \centering
    \begin{picture}(110,85)
    \put(-4,88){\includegraphics[angle=270,width=120\unitlength]
        {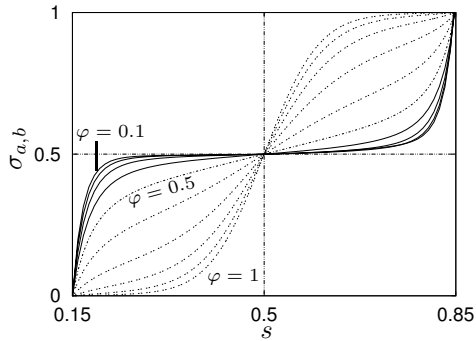}}
      \put(-1,44){\begin{sideways}$\sigma_{a,b}$\end{sideways}}
      \put(60,3){$s$}
      \put(15,52){\scriptsize $\varphi=0.1$}
      \put(28,35.5){\begin{rotate}{15}\scriptsize $\varphi=0.5$\end{rotate}}
      \put(47,17){\scriptsize $\varphi=1$}
      \put(20,44){\line(0,1){7}}
      %@generator gnuplot
      %@script figs/data/collectiveDecision/urnModel/split.plt
      %@data figs/data/collectiveDecision/urnModel/split_*
      %\graphpaper[10](0,0)(110,85)
    \end{picture}
  \caption{\label{fig:split}Splitting probabilities~$\sigma_{a,b}$
    between the two peaks in the steady state probability
    density~$\pi$ for varied positive feedback
    intensity~$\varphi\in\{0.1,0.2,\dots,1\}$.}
\end{figure}

Another statistical property of Markov processes, especially those
with bistable potentials, is the mean first passage time. Of interest
is the mean time to switch from the collective decision for option~$A$
($s=s_1$ with $\Delta B(s_1)=0$) to option~$B$ ($s=s_2$ with $\Delta
B(s_2)=0$) or vice versa due to symmetry. The switching time is of
particular interest in the context of swarm intelligence because it
tells whether a swarm is able to stay for a considerable time in a
given state (e.g., see~\citet{yates09}). In case of collective motion,
the mean switching time describes whether the swarm will be aligned
long enough to cover a considerable distance.

Markov theory allows to calculate the mean first passage time using
the transition matrix~{\bf T} and specifying a target state~$i$ which
is defined to be absorbing ($T_{i,i}=1$), that is, we convert the
system into an absorbing Markov chain. Using the fundamental
matrix~${\bf M}=({\bf I}-{\bf Q})^{-1}$ (with identity matrix ${\bf
  I}$ and ${\bf Q}$ is the transition matrix of transitional states),
a vector of mean first passage times for all transitional states is
obtained by ${\bf Mc}$ with ${\bf c}$ is a column vector of all
1's~\citep{grinstead97}. In addition, an estimate of the mean
switching time can be determined numerically in the urn model by
generating trajectories~$s(t)$ and observing the switching
behavior. This estimate is a lower bound (finite simulation time,
finite number of samples, power-law distribution of first passage
times). A~comparison between theory and simulation along with fitted
functions ($\tau(N)=a_1N^ba_2\exp(cN)$) is shown in
Fig.~\ref{fig:switchTimes}. The switching time scales approximately
exponentially with swarm size~$N$. The fact that
Eq.~\ref{eq:perf:fullEq} is also a good choice in fitting the mean
first passage times can be interpreted as more than mere
coincidence. The mean first passage time may be seen as performance
measure because longer times reflect a more stable swarm. However, for
Eq.~\ref{eq:perf:fullEq} we require~$c<0$ whereas here fitting results
in~$c>0$ which would be interpreted as a positive effect of
interference (cf.~\ref{eq:perf:interference}).

\setlength{\unitlength}{0.005\textwidth}
\begin{figure}
  \centering
    \begin{picture}(110,85)
    \put(0,88){\includegraphics[angle=270,width=120\unitlength]
        {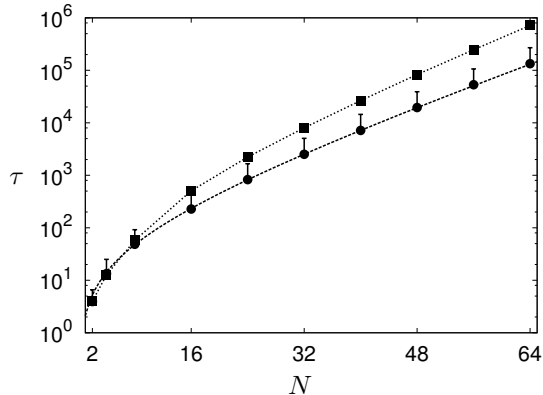}}
      \put(-4,46){$\tau$}
      \put(57,0){$N$}
      %@generator gnuplot
      %@script figs/data/collectiveDecision/urnModel/switchTimes.plt
      %@data figs/data/collectiveDecision/urnModel/switchTimesData.tar.gz
      %\graphpaper[10](0,0)(110,85)
    \end{picture}
  \caption{\label{fig:switchTimes}Mean first passage times~$\tau$ between
    the two collective decisions $s_1$ and $s_2$ with $\Delta
    B(s_1)=\Delta B(s_2)=0$ over swarm size~$N$, squares give values
    according to Markov theory, circles give measurements of the urn
    model, error bars give standard deviations, lines are fitted
    functions $\tau(N)=a_1N^ba_2\exp(cN)$.}
\end{figure}

\subsection{Examples}
\label{collectiveDecisions:examples}

Next we want to compare the data from our urn model
(Fig.~\ref{fig:symBreak:change}) to data from more complex models,
such as the density classification scenario~\citep{hamann07e}. In the
following, we apply the more general definition of $\Delta B$
(Eq.~\ref{eq:DeltaBextended}) but with an invariant payoff~$M(s)=c$
for a scaling constant~$c$ that scales the average change for
different payoffs.

\setlength{\unitlength}{0.0035\textwidth}
\begin{figure}%[h]
  \centering
  \subfigure[\label{fig:densityEst:change}Density classification scenario~\citep{hamann07e}, change of the ratio of red robots for different times during simulation, squares give empirical data (from~\citep{hamann10c}), lines are fitted according to Eq.~\ref{eq:DeltaBextended}.]{
    \begin{picture}(130,90)
      \put(-7,95){\includegraphics[angle=270,width=130\unitlength]
        {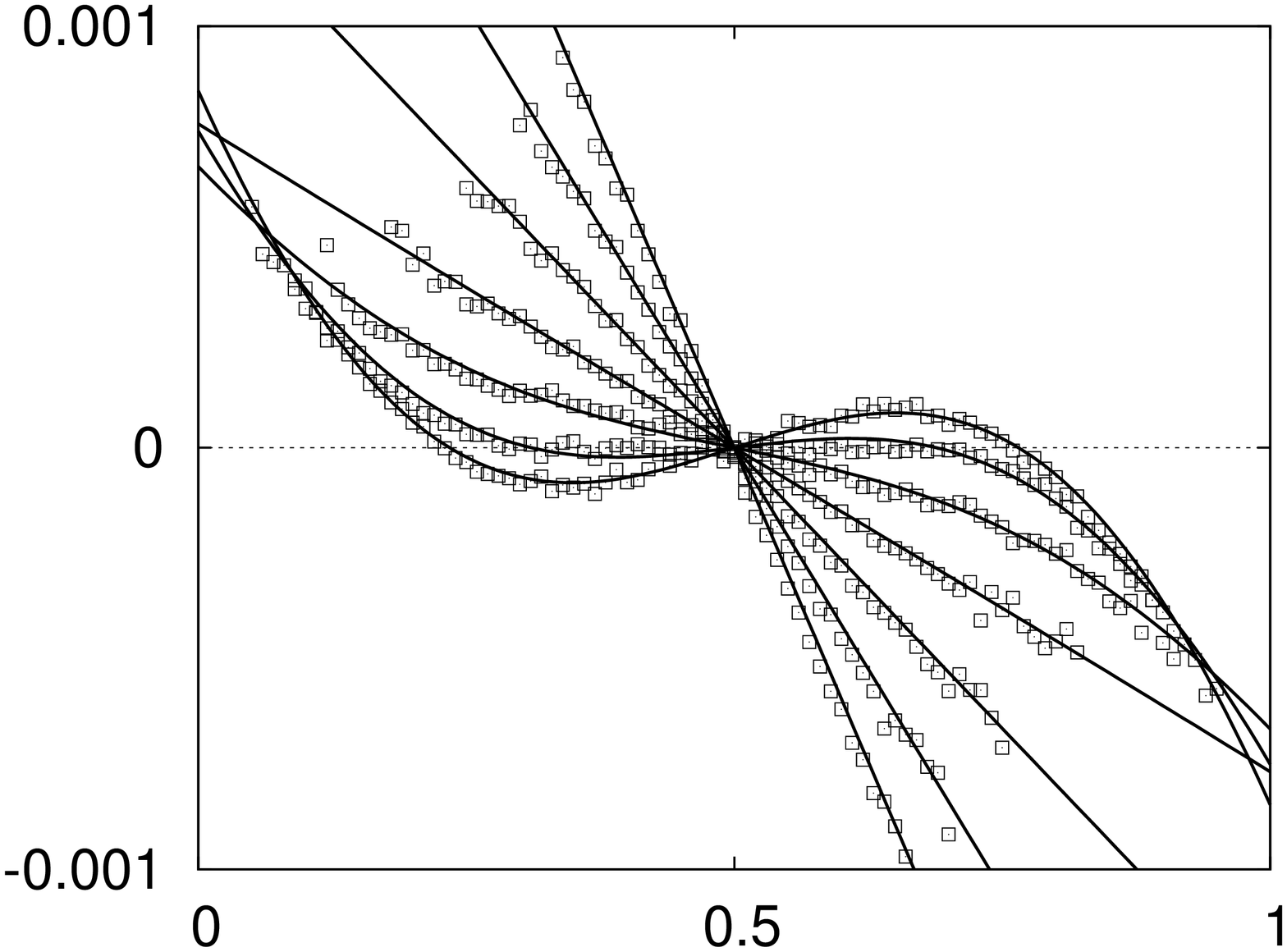}}
      \put(65,1){\small $s$}
      \put(0,39){\begin{sideways}\small $\overline{\Delta s(s)/\Delta t}$\end{sideways}}
      % @generator gnuplot
      % @script figs/data/densityEst.plt
      % @data figs/data/densityEstFracChange200kSamplesNoiseEdited*
      % \graphpaper[10](0,0)(120,90)
    \end{picture}
  }
  \hspace{1.5mm}
  \subfigure[\label{fig:densityEst:posFeedbackFit}Negative exponential function $\varphi(t)=0.786-\exp(-5\times10^{-4}t)$ fitted to feedback intensities obtained from the density classification scenario.]{
    \begin{picture}(130,90)
      \put(-7,95){\includegraphics[angle=270,width=130\unitlength]
        {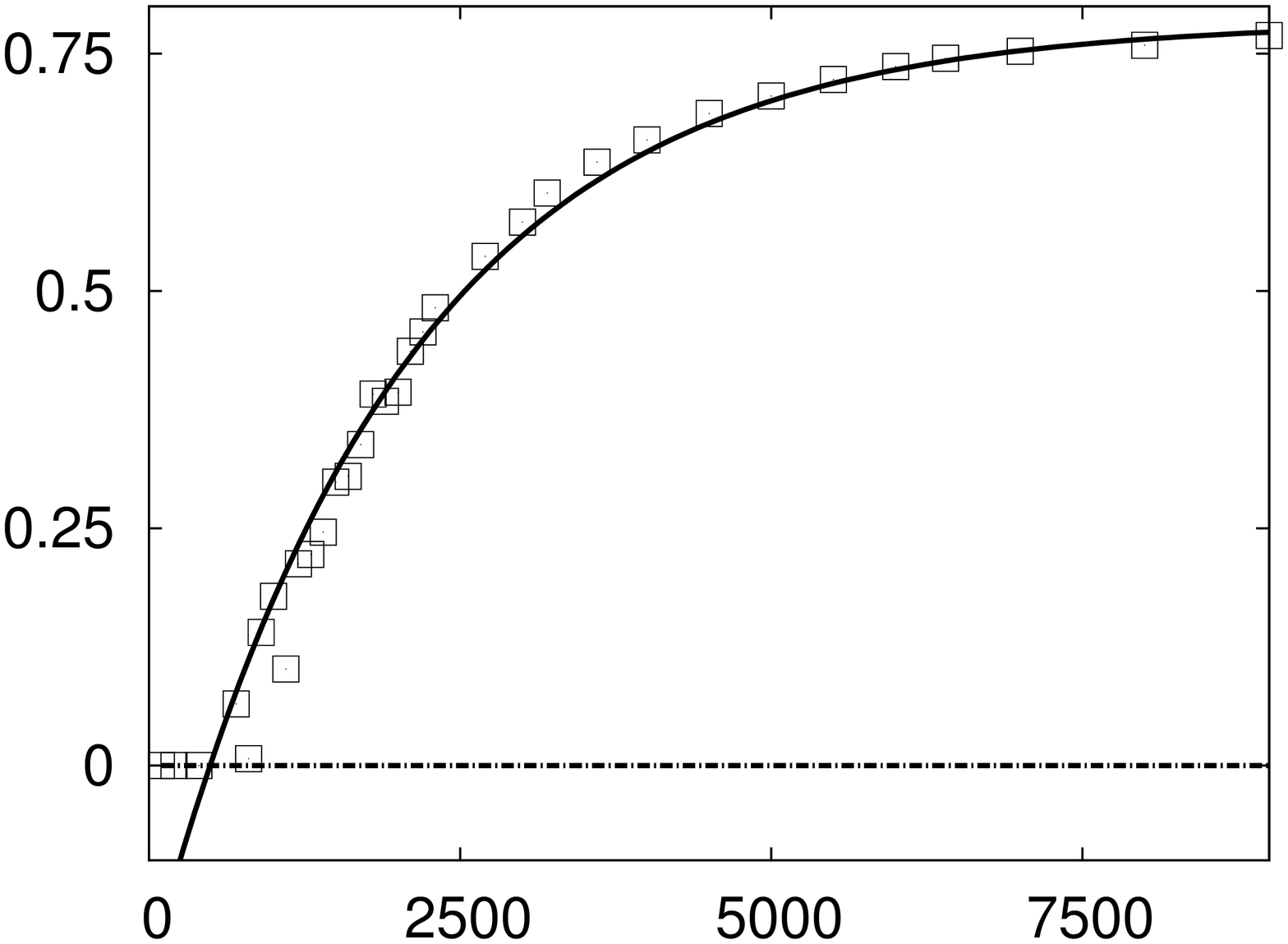}}
      \put(66,1){\small $t$}
      \put(-5,50){\small $\varphi$}
      % @generator gnuplot
      % @script figs/data/densityEst.plt
      % @data figs/data/densityEstFittedPosFeedback
      % \graphpaper[10](0,0)(120,90)
    \end{picture}
  }
  \caption{\label{fig:densityEst}Comparison of model and results from
    the density classification scenario~\citep{hamann07e} and increase
    of positive feedback over time.}
\end{figure}

The density classification scenario~\citep{hamann07e} is about a swarm
of red and green agents moving around randomly. Their only interaction
is constantly keeping track of those agents' colors they bump
into. Once an agent has seen five agents of either color it changes
its own color to that it has encountered most. Here, $s$ gives the
ratio of red agents. The name of this scenario is due to the idea that
the swarm should converge to that color that was initially superior in
numbers. It turns out that the averaged change~$\overline{\Delta
  s(s)/\Delta t}$ (see Fig.~\ref{fig:densityEst:change}) starts with a
curve similar to that of~$\varphi=0$ in Fig.~\ref{fig:symBreak:change}
and then converges to a curve that is similar to that
of~$\varphi=0.75$. That is, $\overline{\Delta s(s)/\Delta t}$~is
time-variant and, for example, the above mentioned Markov model would
have to be extended to a time-inhomogeneous Markov model to achieve a
better correlation with the measurements. Early in the simulation
there is mostly negative feedback forcing values close to~$s=0.5$. The
negative feedback decreases with increasing time, which results finally
in positive feedback for $s\in[0.23,0.77]$. Comparing
Fig.~\ref{fig:symBreak:change} to Fig.~\ref{fig:densityEst:change}
indicates a good qualitative agreement between our urn model and the
density classification scenario. Given that the curves in
Fig.~\ref{fig:densityEst:change} converge over time to the final shape,
which is resembled by our model for increasing~$\varphi$ in
Fig.~\ref{fig:symBreak:change}, one can say that positive feedback
builds up slowly over time in the density classification scenario. By
fitting Eq.~\ref{eq:DeltaBextended} to the data shown in
Fig.~\ref{fig:densityEst:change} we get estimates for the feedback
intensity~$\varphi$. From the earliest and steepest line to the latest
and only curve with positive slope in $s=0.5$ we get values of
$\varphi\in [0,0,0,0.007,0.304,0.603]$ for times~$t\in
[100,200,400,800,1600,3200]$. By continuing this fitting for
additional data not shown in Fig.~\ref{fig:densityEst:change}, we are
able to investigate the temporal evolution of feedback
intensity~$\varphi$ according to our model. In
Fig.~\ref{fig:densityEst:posFeedbackFit}, the data points of feedback
intensity~$\varphi$ obtained by fitting are shown along with a
function that was fitted to the data. This result supports the
assumption of a `negative exponential' increase over time
($1-\exp(-t)$) of positive feedback in this system as already stated
in~\citep{hamann10c}.

\setlength{\unitlength}{0.0035\textwidth}
\begin{figure}%[h]
  \centering
  \subfigure[\label{fig:measuredFB:fb}Positive feedback probability as measured in the simulation according to Eq.~\ref{eq:redeterminFB} (squares) and fitted function of Eq.~\ref{eq:polynomFB} (line).]{
    \begin{picture}(120,90)
      \put(0,95){\includegraphics[angle=270,width=130\unitlength]
        {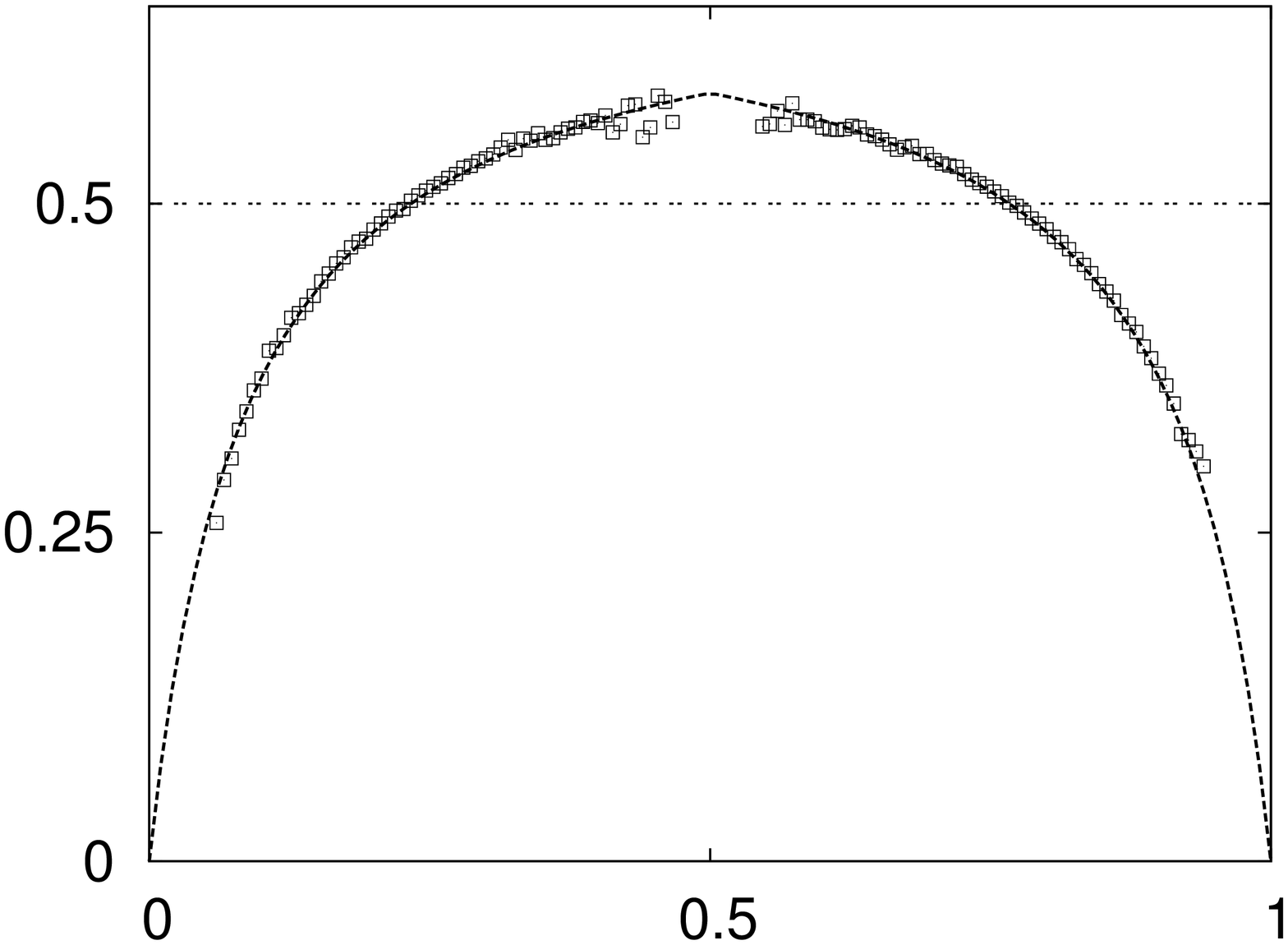}}
      \put(70,1){\small $s$}
      \put(0,49){\small $P$}
      % @generator gnuplot
      % @script figs/data/collectiveDecision/densityEst/densityEst.plt
      % @data figs/data/collectiveDecision/densityEst/measuredPositiveFeedback
      % \graphpaper[10](0,0)(120,90)
    \end{picture}
  }
  \hspace{1.5mm}
  \subfigure[\label{fig:measuredFB:change}Predicted average change based on the measurements shown in (a) and Eq.~\ref{eq:DeltaBextended} (line) compared to direct measurements in simulation (squares).]{
    \begin{picture}(130,90)
      \put(0,95){\includegraphics[angle=270,width=130\unitlength]
        {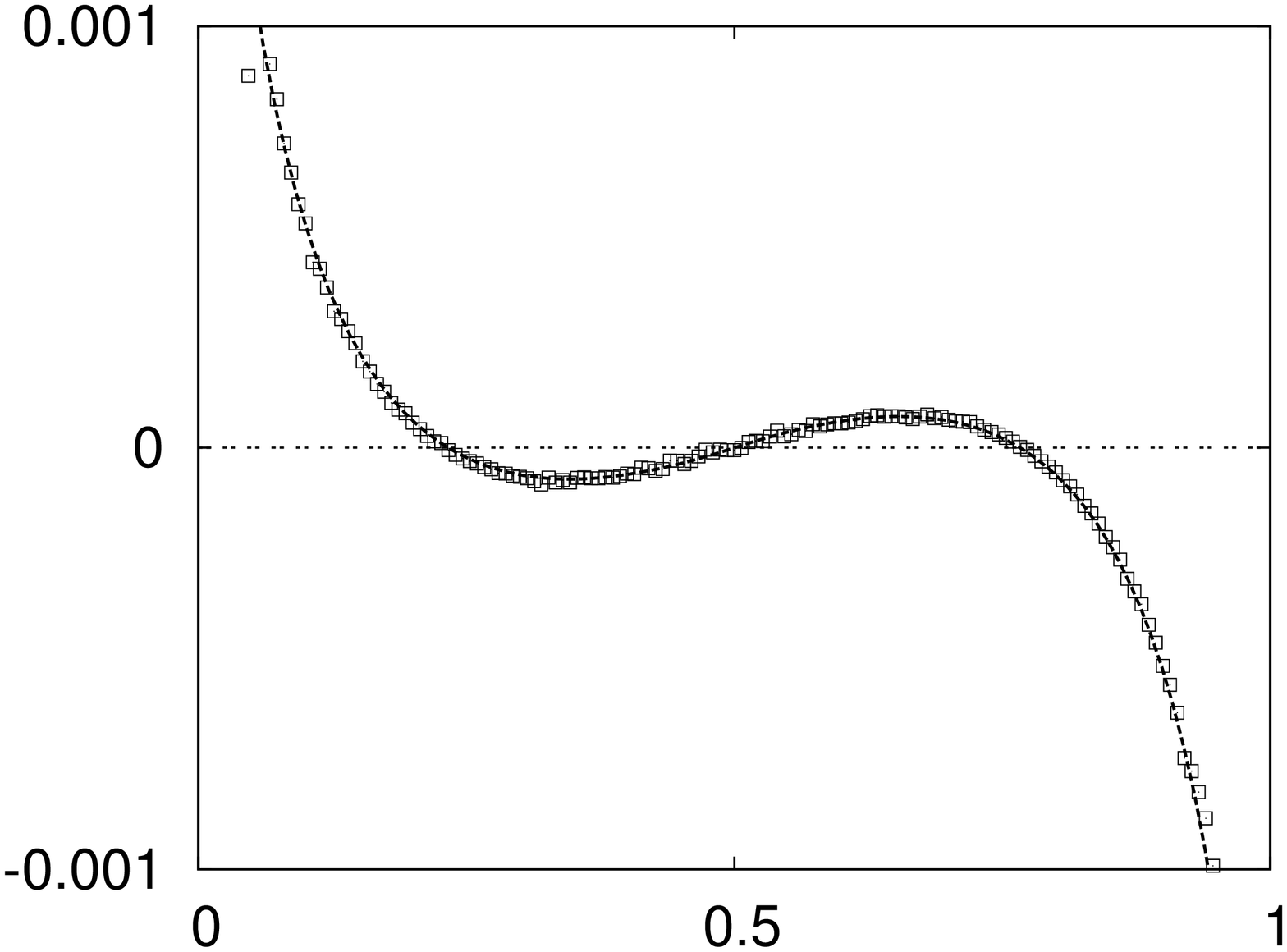}}
      \put(72,1){\small $s$}
      \put(0,37){\small \begin{sideways}${\Delta s(s)/\Delta t}$\end{sideways}}
      % @generator gnuplot
      % @script figs/data/collectiveDecision/densityEst/densityEst.plt
      % @data figs/data/collectiveDecision/densityEst/newChangeHisto
      %\graphpaper[10](0,0)(120,90)
    \end{picture}
  }
  \subfigure[\label{fig:measuredFB:steadyState}Predicted steady state obtained by using the measurements shown in (a) and by applying Eqs.~\ref{eq:DeltaBextended}, \ref{eq:transitionMatrix}, and~\ref{eq:eigenvectors} (line); direct measurement in simulation (squares).]{
    \begin{picture}(130,90)
      \put(0,95){\includegraphics[angle=270,width=130\unitlength]
        {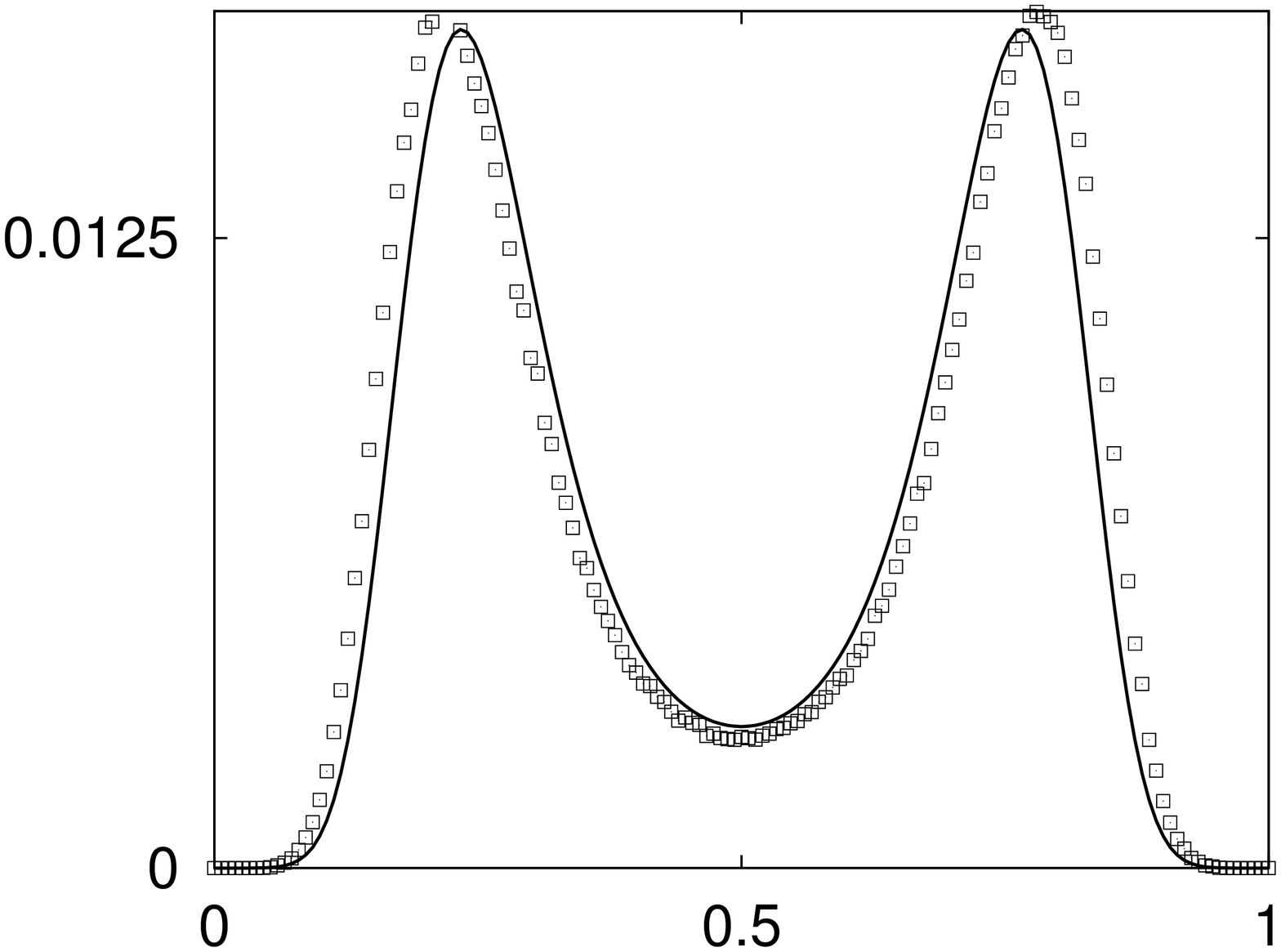}}
      \put(73,1){\small $s$}
      \put(-2,45){\begin{sideways}\small $\pi(s)$\end{sideways}}
      % @generator gnuplot
      % @script figs/data/collectiveDecision/densityEst/densityEst.plt
      % @data figs/data/collectiveDecision/densityEst/markovSteadyState
      %\graphpaper[10](0,0)(120,90)
    \end{picture}
  }
  \caption{\label{fig:measuredFB}Measured positive feedback
    probability, predicted and measured average change~$\Delta s$, and
    predicted and measured steady state for the density classification
    scenario.}
\end{figure}

The positive feedback probability~$P(s)$ can also be measured via the
observed decision revisions according to
Eq.~\ref{eq:redeterminFB}. This was done for the density
classification scenario; the results are given in
Fig.~\ref{fig:measuredFB:fb}, which shows the measured positive
feedback probability~$P(s)$ for time $t=7900$. The data was fitted
using %a function that is a translated Wigner semicircle distribution
%% \begin{equation}
%%   \label{eq:semiCircle}
%%   P(s)=\frac{2a}{\pi b^2}\sqrt{b^2-(2s-1)^2},
%% \end{equation}
%for free parameters $a$ and $b$, 
%a polynom of degree four
the following axially symmetric function
\begin{equation}
  \label{eq:polynomFB}
%  P(s)=c_4s^4+c_3s^3+c_2s^2+c_1s+c_0,
  P(s)=
\begin{cases}
  c_1\left(1-\frac{1}{1+c_2s}\right),     & \text{ for }s\le 0.5\\
  c_1\left(1-\frac{1}{1+c_2(1-s)}\right), & \text{ else}
\end{cases} \enspace ,
\end{equation}
for constants $c_1$ and $c_2$, which turns out to be a better fit here
than the sine function. Once the function is fitted, it can be used to
calculate the expected average change~$\Delta s$ following
Eq.~\ref{eq:DeltaBextended} (only scaling constant~$c$ needs to be
fitted). As seen in Fig.~\ref{fig:measuredFB:change}, we obtain a good
fit (root mean squared error of~$<9\times 10^{-3}$). Indeed, our experience
shows that the procedure of fitting~$P(s)$ instead of~$\Delta s$
directly is much more accurate and needs fewer samples. Especially, it
places the zeros~$\Delta s=0$ more accurate, which are important to
predict the correct steady state of the system. Even with just
100~samples, good fits were obtained, which indicates that this model
could be applied to data from natural swarm systems. The prediction of
the steady state based on the measured positive feedback probability
following Eqs.~\ref{eq:DeltaBextended}, \ref{eq:transitionMatrix},
and~\ref{eq:eigenvectors} is shown in
Fig.~\ref{fig:measuredFB:steadyState}. In comparison to the
measurements from simulation, this prediction shows a reasonable
agreement.

\setlength{\unitlength}{0.004\textwidth}
\begin{figure}%[h]
  \centering
  \begin{picture}(130,90)
    \put(-7,95){\includegraphics[angle=270,width=130\unitlength]
      {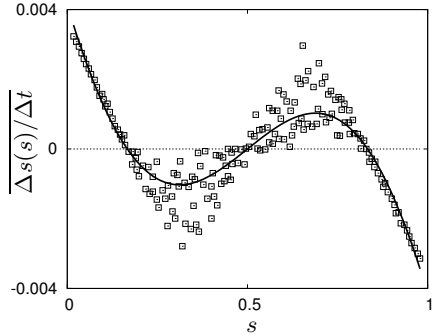}}
    \put(65,1){\small $s$}
    \put(0,39){\begin{sideways}\small $\overline{\Delta s(s)/\Delta t}$\end{sideways}}
    % @generator gnuplot
    % @script figs/data/yates.plt
    % @data figs/data/yates
    % \graphpaper[10](0,0)(120,90)
  \end{picture}
  \caption{\label{fig:densityEst:yates}Model fitted to data from
    Fig.~3B of \citet{yates09} (local model of swarm alignment in
    locusts) by $\varphi=1$ (and $c=4.134\times10^{-3}$); data scaled
    to $s\in{[0,1]}$.}
\end{figure}

Other examples showing similarities to the $\varphi=0.75$-graph in
Fig.~\ref{fig:symBreak:change} are Figs.~2B and~3B in \citet{yates09}
which show the drift coefficient dependent on the current alignment of
a swarm (average velocity). While the data obtained from experiments
with locusts, Fig.~2B in~\citep{yates09}, is too noisy, we use the
data from their model, Fig.~3B in~\citep{yates09}, to fit our
model. The result is shown in Fig.~\ref{fig:densityEst:yates}. We
obtain a maximal positive feedback of~$\varphi=1$.

\section{Discussion}

In this paper, we have reported two abstract models of swarms with
high generality due to our long-term objective of creating a swarm
calculus. The first model focuses on the dependency of swarm
performance on swarm density by separating the system into two parts:
cooperation and interference. It explains that an optimal or critical
swarm density exists at which the peak performance is reached. With
the second model we describe the dynamics of collective decision
processes with focus on the existence and intensity of feedback. It
gives an explanation of how the typical cubic functions of decision
revision emerge by an increase of positive feedback over time.

The first model is simple and the existence of optimal swarm densities
is a well-known fact. However, to the authors knowledge, no similar
model combined with a validation by fitting the model to data from
diverse swarm applications was reported before except for the
hypotheses stated by \citet{ostergaard01}. Despite its simplicity, the
model has the capability to give predictions of swarm performance,
especially, if the available data, to which it is fitted, includes an
interval around the optimal density. That way this model might serve
as a swarm calculus of swarm performance. In addition, we want to draw
attention to the problem of masking special density-dependent
properties by only investigating the mean performance. The example
shown in Fig.~\ref{fig:taxisHist2d} documents the existence of phases
in swarm systems.

The second model is abstract as well but has a higher complexity and
is more conclusive as it allows for mathematical
derivations. Based on this urn model for positive feedback decision
processes, the emerging cubic function of decision revisions can be
derived (see Eq.~\ref{eq:DeltaB}). Hence, we generate the function of
decision revision based on our urn model, which allows for an
interpretation of how the function emerges while, for example,
in~\citep{yates09} this function is measured in a local model. Our
model of collective decisions might qualify as a part of swarm
calculus because those decision revision functions seem to be a
general phenomenon in swarms.

This model can also be used to predict probability density functions
of steady states in swarm systems. The workflow of measuring the
positive feedback probability~$P(s)$, fitting a function, using this
function to calculate the expected average change~$\Delta s$, which
can in turn be used to predict the expected probability density
function of the steady state (eigenvectors of transition matrix), is
accurate with comparatively small sample numbers.

A model of notable similarity was published in the context of
`sociophysics' \citep{galam04}. It is based on the assumption that
subgroups form in collective decisions within which a majority rule
determines the subgroup's decision. The addition of contrarians, that
is voters always voting against the current majority decision,
generates dynamics that are similar to the reported observations in
noisy swarms. Galam's model is, however, focused on constant sizes of
these subgroups and their combinatorics while in swarm intelligence
these groups and their sizes are dynamic.

A result of interest is also the particular function of the positive
feedback increase over time ($1-\exp(-t)$) in the density
classification task (see Fig.~\ref{fig:densityEst:posFeedbackFit}). It
is to be noted that this increase seems to be independent from
respective values of the current consensus~$s$. Furthermore, extreme
values, such as $s \approx 1$ or $s \approx 0$, are not observed. An
in-depth analysis of the underlying processes is beyond this paper but
we want to present two ideas. First, the final saturation phase
($\lim_{t\rightarrow \infty}\varphi=0.8$) is most likely caused by
explicit noise in the simulation. The agent--agent recognition rate
was set to 0.8 which keeps $P(s,\varphi)$ small. Second, the initial
fast increase of~$\varphi$ (after a transient which might also be
caused by the simulation because agents revise their color only after
a minimum of five agent--agent encounters) might be caused by locally
emerging sub-groups of homogeneous color within small areas that
generate `islands' of early positive feedback. These properties might,
however, be highly stochastic and difficult to measure. Time-variant
positive feedback was also observed in BEECLUST-controlled swarms as
reported before~\citep{hamann10a}. Hence, a feedback system as given
in Fig.~\ref{fig:feedbackDiagram} seems to be a rather common
situation in swarm systems. $A$~and $B$ are properties of a swarm
system that form a feedback system (e.g., amount of pheromone and
number of recruited ants). $C$~is a third swarm property, which is
subject to a time-variant process and which influences the feedback of
$B$ on~$A$. In terms of the above urn model we can mimic this
situation by saying $A$~in Fig.~\ref{fig:feedbackDiagram} is the
number of blue marbles (w.l.o.g.), $B$~is the probability of drawing a
blue marble, $P$~is the probability of positive feedback (i.e., this
edge can also negatively influence~$A$), and $C$~is an unspecified
state variable that increases positive feedback ($\varphi$) over time
and is influenced by an additional, unknown process. This triggers the
question of what~$C$ can be and how it influences the feedback process
independent of the current swarm consensus~$s$.

\setlength{\unitlength}{0.003\textwidth}
\begin{figure}%[h!]
  \centering
  \begin{picture}(125,85)
    \put(0,0){\includegraphics[angle=0,width=130\unitlength]
      {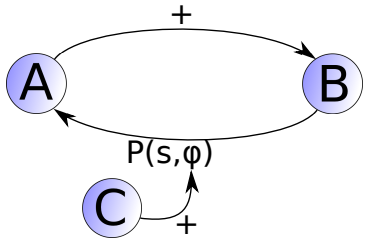}}
    %@generator inkscape
    %@data figs/data/feedbackDiagram.svg
    %\graphpaper[10](0,0)(125,85)
  \end{picture}
  \caption{\label{fig:feedbackDiagram}Time-variant feedback system; here for increasing probability of positive feedback.}
\end{figure}

As seen in Fig.~\ref{fig:densityEst:yates}, we get maximally positive
feedback~$\varphi=1$ for the data of~\citet{yates09} which has the
effect that states of low alignment ($s\approx 0.5$) are left as fast
as possible. This reinforces the findings about the diffusion
coefficient reported by \citet{yates09}. A~major feature of this
self-organizing swarm seems to be the minimization of times in states
of low alignment (Yates et al.: ``A~higher diffusion coefficient at
lower alignments suggests that the locusts `prefer' to be in a highly
aligned state'').

\section{Conclusion}

On of the main results reported in this paper is that generally
applicable swarm models, that have simple preconditions, exist. The
application of the swarm performance model necessitates only a concept
of swarm density and the application of the collective decisions model
necessitates only a consensus variable of a binary decision. Although
both models are simple, they have enough explanatory power to give
insights into swarm processes such as the interplay of cooperation and
interference and the installation of positive feedback.

The two presented models illustrate the methodology that can be
applied to find more models and to extend swarm calculus. The
methodology is characterized by a combination of a heuristic approach
and a simple mathematical formalism. While the empirical part
establishes a direct connection to applications, the formalism allows
the integration of superior mathematical methods such as Markov theory
and linear algebra. That way there is reason for hope that similar
models might be found, for example, for swarm systems showing
aggregation, flocking, synchronization, or self-assembly. The main
benefit of such models might be general insights in group behavior and
swarms but also direct applications could be possible. It could be
possible to implement variants of these models on swarm robots if the
global knowledge necessary for this kind of models can be substituted
by local samplings. The models could also be used as components and
several such components could be combined to form specialized,
sophisticated models. The presented models could be combined to model
a swarm showing collective decision--making, such as a
BEECLUST-controlled swarm as pointed out above. Once a comprehensive
set of such models has been collected, research on swarms could also
be guided by formalisms supplemental to empirical research. Hence,
we contend that it is possible to generate a set of models and methods
of general applicability for swarm science, that is, to create a swarm
calculus.

\input{appendixFitting}

\end{document}

%% file: appendixFitting.tex
\appendix
 
\section{Details on curve fitting}

All curve fitting was done with an implementation of the nonlinear
least-squares Marquardt-Levenberg
algorithm~\citep{levenberg44,marquardt63} using gnuplot~4.6
patchlevel~1 (2012-09-26)\footnote{see
  \url{http://www.gnuplot.info/}}.
%http://www.gnuplot.info/}.

\subsection{Foraging in a group of robots}
% Fig 2b
The data for the curve fitted in Fig.~\ref{fig:fitted:lerman} is shown
in Tab.~\ref{tab:fit:foraging}.

      %@script figs/data/performance/lerman/lermanDiagram.plt
      %@data figs/data/performance/lerman/lermanDiagramData

\begin{table}[h]
  \begin{center}
    \begin{tabular}{r|c}
      \hline
      function & $P(N)=a_1N^b a_2\exp(cN)$ \\
      degrees of freedom & 52 \\
      root mean square of residuals & 0.000146389 \\
      \hline
    \end{tabular}
    \begin{tabular}{c|c|cc}
      parameter & value & asymptotic standard error & \\
      \hline
$a_1a_2$          & 0.00248537  &     +/- 4.499e-05 &   (1.81\%)\\
$b$               & 1.23745     &     +/- 0.01969   &   (1.591\%)\\
$c$               & -0.199589   &     +/- 0.002932  &   (1.469\%)\\
      \hline
    \end{tabular}
  \end{center}
  \caption{\label{tab:fit:foraging}Fitting data, foraging in a group
    of robots, Fig.~\ref{fig:fitted:lerman}.}
\end{table}

\subsection{Collective decision making based on BEECLUST}
% Fig 2c
The data for the curve fitted in Fig.~\ref{fig:fitted:symBreakScan} is
shown in Tab.~\ref{tab:fit:collectiveDecisionBeeclust}.
 
      %@script figs/data/performance/symBreak/symBreakScan.plt
      %@data figs/data/performance/symBreak/symBreakScan*

\begin{table}[h]
  \begin{center}
    \begin{tabular}{r|c}
      \hline
      function & $P(N)=a_1N^b a_2\exp(cN)$ \\
      degrees of freedom & 22 \\
      root mean square of residuals & 0.0515291 \\
      \hline
    \end{tabular}
    \begin{tabular}{c|c|cc}
      parameter & value & asymptotic standard error & \\
      \hline
$a_1a_2$          & 0.158797     &    +/- 0.02234   &   (14.07\%)\\
$b$               & 0.772042     &    +/- 0.06951   &   (9.003\%)\\
$c$               & -0.0386915   &    +/- 0.002867  &   (7.409\%)\\
      \hline
    \end{tabular}
  \end{center}
  \caption{\label{tab:fit:collectiveDecisionBeeclust}Fitting data,
    collective decision making based on BEECLUST,
    Fig.~\ref{fig:fitted:symBreakScan}.}
\end{table}

\subsection{Aggregation in tree-like structures and reduction to shortest path}
% Fig 2d
The data for the curve fitted in Fig.~\ref{fig:fitted:diplomarbeit} is
shown in Tab.~\ref{tab:fit:diplomarbeit}. In this case weighted
fitting was used (values $\rho_1=0.00524$ and $\rho_2=0.00598$ were
weighted ten times higher than other values) to enforce the
limit~$P<1$.

      %@script figs/data/performance/diplomarbeit/diplomarbeit.plt
      %@data figs/data/performance/diplomarbeit/diplomarbeitData

\begin{table}[h]
  \begin{center}
    \begin{tabular}{r|c}
      \hline
      function & $P(\rho)=a_1\rho^b a_2\exp(c\rho)$ \\
      degrees of freedom & 21 \\
      root mean square of residuals & 0.0924653 \\
      \hline
    \end{tabular}
    \begin{tabular}{c|c|cc}
      parameter & value & asymptotic standard error & \\
      \hline
$a_1a_2$         &  114.55      &     +/- 49.11     &   (42.87\%)\\
$b$              &  0.836024    &     +/- 0.07586   &   (9.074\%)\\
$c$              &  -89.9857    &     +/- 6.985     &   (7.763\%)\\
      \hline
    \end{tabular}
  \end{center}
  \caption{\label{tab:fit:diplomarbeit}Fitting data, aggregation in tree-like structures and reduction to shortest path, Fig.~\ref{fig:fitted:diplomarbeit}.}
\end{table}

\subsection{Emergent--taxis behavior}
\label{fit:emergentTaxis}
% Fig 3a
The data for the curve fitted in Fig.~\ref{fig:taxis:fitted} is shown
in Tab.~\ref{tab:fit:emergentTaxis}. Fitting was done in two
steps. First, the interference function~$I(N)$ was fitted. Second, the
performance function~$P(N)$ was fitted while keeping the parameters
$a_2$ and $c$ fixed.

      %@script figs/data/performance/taxis/taxisSuccessRate.plt
      %@data figs/data/performance/taxis/taxisSuccessRate*
\begin{table}[h!]
  \begin{center}
    %first fit:\\
    \begin{tabular}{r|c}
      \hline
      function & $I(N)=a_2\exp(cN)+d$ \\
      degrees of freedom & 48 \\
      root mean square of residuals & 0.00479438 \\
      \hline
    \end{tabular}
    \begin{tabular}{c|c|cc}
      parameter & value & asymptotic standard error & \\
      \hline
$a_2$             & 0.213822     &    +/- 0.007214  &   (3.374\%)\\
$c$               & -0.182333    &    +/- 0.007664  &   (4.203\%)\\
$d$               & 0.0750781    &    +/- 0.0008863 &   (1.181\%)\\
      \hline
    \end{tabular}
    %second fit:\\
    \begin{tabular}{r|c}
      \hline
      function & $P(N)=a_1N^b a_2\exp(cN)$ \\
      degrees of freedom &  41\\
      root mean square of residuals & 0.0403196 \\
      \hline
    \end{tabular}
    \begin{tabular}{c|c|cc}
      parameter & value & asymptotic standard error & \\
      \hline
$a_1$             & 0.0106104   &     +/- 0.0009767 &   (9.205\%)\\
$b$               & 3.23718     &     +/- 0.03055   &   (0.9438\%)\\
      \hline
    \end{tabular}
  \end{center}
  \caption{\label{tab:fit:emergentTaxis}Fitting data, emergent--taxis
    behavior, Fig.~\ref{fig:taxis:fitted}.}
\end{table}

\subsection{Emergent--taxis behavior, narrow fit}
% Fig 3b
The data for the curve fitted in Fig.~\ref{fig:taxis:narrowFit} is
shown in Tab.~\ref{tab:fit:emergentTaxisNarrowFit}. The
parameters~$a_2$ and~$c$ of the interference function~$I(N)$ as
obtained in~\ref{fit:emergentTaxis} were reused. The performance
function~$P(N)$ was fitted within the narrow interval of $N\in
[20,22]$ while keeping the parameters $a_2$ and $c$ fixed.

      %@script figs/data/performance/taxis/taxisSuccessRate.plt
      %@data figs/data/performance/taxis/taxisSuccessRate*
\begin{table}[h!]
  \begin{center}
     \begin{tabular}{r|c}
      \hline
      function & $P(N)=a_1N^b a_2\exp(cN)$ \\
      degrees of freedom &  1\\
      root mean square of residuals & 0.0180345 \\
      \hline
    \end{tabular}
    \begin{tabular}{c|c|cc}
      parameter & value & asymptotic standard error & \\
      \hline
$a_1$             & 0.00660836    &   +/- 0.005772   &  (87.34\%)\\
$b$               & 3.38946       &   +/- 0.2856     &  (8.425\%)\\
      \hline
    \end{tabular}
  \end{center}
  \caption{\label{tab:fit:emergentTaxisNarrowFit}Fitting data, emergent--taxis
    behavior, narrow fit, Fig.~\ref{fig:taxis:narrowFit}.}
\end{table}

\subsection{Mean first passage times}
% Fig 7
The data for the curve fitted in Fig.~\ref{fig:switchTimes} is shown
in Tab.~\ref{tab:fit:switchTimes}. Weighted fitting was applied based
on the measured standard deviation and weights scaled by
$\sqrt{\tau^{\text{theor}}}$ respectively.

      %@script figs/data/collectiveDecision/urnModel/switchTimes.plt
      %@data figs/data/collectiveDecision/urnModel/switchTimesData.tar.gz
\begin{table}[h]
  \begin{center}
    \begin{tabular}{r|c}
      \hline
      function measured $\tau$ & $\tau^{\text{meas}}(N)=a_1^{\text{meas}}N^{b^{\text{meas}}} a_2^{\text{meas}}\exp(c^{\text{meas}}N)$ \\
      function theoretical $\tau$ & $\tau^{\text{theor}}(N)=a_1^{\text{theor}}N^{b^{\text{theor}}} a_2^{\text{theor}}\exp(c^{\text{theor}}N)$ \\
      degrees of freedom &  7\\
      rms of residuals ($\tau^{\text{meas}}$) & 0.0613924 \\
      rms of residuals ($\tau^{\text{theor}}$) & 1.35583 \\
      \hline
    \end{tabular}
    \begin{tabular}{c|c|cc}
      parameter & value & asymptotic standard error & \\
      \hline
$a_1^{\text{meas}}a_2^{\text{meas}}$ & 1.36333   &       +/- 0.07285  &   (5.343\%)\\
$b^{\text{meas}}$               & 1.31916    &      +/- 0.03673   &   (2.784\%)\\
$c^{\text{meas}}$               & 0.0933643  &      +/- 0.002197  &   (2.353\%)\\
$a_1^{\text{theor}}a_2^{\text{theor}}$ & 1.31234  &        +/- 0.2235 &   (17.03\%)\\
$b^{\text{theor}}$              & 1.52047     &     +/- 0.05814   &   (3.824\%)\\
$c^{\text{theor}}$              & 0.107615    &     +/- 0.001153  &   (1.072\%)\\
      \hline
    \end{tabular}
  \end{center}
  \caption{\label{tab:fit:switchTimes}Fitting data, mean first passage times, Fig.~\ref{fig:switchTimes}.}
\end{table}

\subsection{Density classification}
% Fig 8a
The data for the curve fitted in Fig.~\ref{fig:densityEst:change} is
shown in Tab.~\ref{tab:fit:densityClassification}. For
times~$t\in\{100,200,400\}$, we set~$\varphi=0$ as otherwise the
fitting would result in $\varphi<0$.

      % @script figs/data/densityEst.plt
      % @data figs/data/densityEstFracChange200kSamplesNoiseEdited*
\begin{table}[h]
  \begin{center}
    \begin{tabular}{r|c|c}
      \hline
      functions & $P(s,\varphi) = \varphi\sin(\pi s)$  \\
               &  $M(s) = c_2$ ($c_1=0$)   \\
               &  $\Delta B(s)=4M(s)(P(s,\varphi)-0.5)(s-0.5)$   \\
      \hline
      time & degrees of freedom   \\
      \hline
      100 & 32 \\
      200 & 40 \\
      400 & 50 \\
      800 & 63 \\
      1600 & 75 \\
      3200 & 81 \\
      6400 & 89 \\
      \hline
      time & root mean square of residuals   \\
      \hline
      100 & 3.29349e-05 \\
      200 & 3.15751e-05 \\
      400 & 2.35511e-05 \\
      800 & 1.94473e-05 \\
      1600 & 1.99464e-05 \\
      3200 & 2.37314e-05 \\
      6400 & 1.99628e-05 \\
      \hline
    \end{tabular}
    \begin{tabular}{c|c|c|cc}
      time & parameter & value & asymptotic standard error & \\
      \hline
      100 & $c_2$         & 0.00297812    &   +/- 3.011e-05 &   (1.011\%) \\
      200 & $c_2$         & 0.00209906    &   +/- 2.084e-05 &   (0.9927\%)\\
      400 & $c_2$         & 0.00133093    &   +/- 1.12e-05  &   (0.8417\%)\\
      800 & $c_2$         & 0.000768213   &   +/- 3.729e-05 &   (4.854\%)\\
      800 & $\varphi$     & 0.00719126    &   +/- 0.0335    &   (465.9\%)\\
      1600 & $c_2$        & 0.000666737   &   +/- 1.904e-05 &   (2.856\%)\\
      1600 & $\varphi$    & 0.304734      &   +/- 0.0148    &   (4.858\%) \\
      3200 & $c_2$        & 0.00075085    &   +/- 1.642e-05 &   (2.186\%) \\
      3200 & $\varphi$    & 0.603136      &   +/- 0.007976  &   (1.322\%)\\
      6400 & $c_2$        & 0.000846385   &   +/- 9.191e-06 &   (1.086\%) \\
      6400 & $\varphi$    & 0.744183     &    +/- 0.004884  &   (0.6563\%) \\
      \hline
    \end{tabular}
  \end{center}
  \caption{\label{tab:fit:densityClassification}Fitting data, density classification, Fig.~\ref{fig:densityEst:change}.}
\end{table}

\subsection{Feedback intensities}
% Fig 8b
The data for the curve fitted in
Fig.~\ref{fig:densityEst:posFeedbackFit} is shown in
Tab.~\ref{tab:fit:fbIntensities}. Weighted fitting was applied with
zero-weight for data points of $t<700$, which means we ignore the
initial values of~$\varphi(t)=0$. Data points of $t\ge 3000$ had
double the weight than values of $700\le t<3000$.

      % @script figs/data/densityEst.plt
      % @data figs/data/densityEstFittedPosFeedback
\begin{table}[h]
  \begin{center}
    \begin{tabular}{r|c}
      \hline
      function & $\varphi(t)=a-\exp(bt)$ \\
      degrees of freedom & 28 \\
      root mean square of residuals & 0.0173061 \\
      \hline
    \end{tabular}
    \begin{tabular}{c|c|cc}
      parameter & value & asymptotic standard error & \\
      \hline
$a$               & -0.000495857  &   +/- 2.064e-05 &   (4.161\%)\\
$b$               & -0.215755     &   +/- 0.01069   &   (4.956\%)\\
      \hline
    \end{tabular}
  \end{center}
  \caption{\label{tab:fit:fbIntensities}Fitting data, feedback intensities, Fig.~\ref{fig:densityEst:posFeedbackFit}.}
\end{table}

\subsection{Positive feedback probability}
% Fig 9a
The data for the curve fitted in Fig.~\ref{fig:measuredFB:fb} is shown
in Tab.~\ref{tab:fit:posFBprob}.

      % @script figs/data/collectiveDecision/densityEst/densityEst.plt
      % @data figs/data/collectiveDecision/densityEst/measuredPositiveFeedback
\begin{table}[h]
  \begin{center}
    \begin{tabular}{r|c}
      \hline
      function & $  P(s)=
\begin{cases}
  c_1\left(1-\frac{1}{1+c_2s}\right),     & s\le 0.5\\
  c_1\left(1-\frac{1}{1+c_2(1-s)}\right), & \text{ else}
\end{cases}$ \\
      degrees of freedom & 120 \\
      root mean square of residuals & 0.00562933 \\
      \hline
    \end{tabular}
    \begin{tabular}{c|c|cc}
      parameter & value & asymptotic standard error & \\
      \hline
$c_1$               & 0.679526   &      +/- 0.001996  &   (0.2938\%)\\
$c_2$               & 11.9802    &      +/- 0.1334    &   (1.113\%)\\
      \hline
    \end{tabular}
  \end{center}
  \caption{\label{tab:fit:posFBprob}Fitting data, positive feedback probability., Fig.~\ref{fig:measuredFB:fb}.}
\end{table}

\subsection{Swarm alignment in locusts}
% Fig 10
The data for the curve fitted in Fig.~\ref{fig:densityEst:yates} is
shown in Tab.~\ref{tab:fit:locusts}. We set~$\varphi=1$ as otherwise
the fitting would result in $\varphi>1$. Weighted fitting was applied
values of~$s<0.17$ and $s>0.83$ had double weight than values of
$0.17\le s \le 0.83$.

    % @script figs/data/yates.plt
    % @data figs/data/yates
\begin{table}[h]
  \begin{center}
    \begin{tabular}{r|c}
      \hline
      functions & $P(s,\varphi) = \varphi\sin(\pi s)$  \\
               &  $M(s) = c_2$ ($c_1=0$)   \\
               &  $\Delta B(s)=4M(s)(P(s,\varphi)-0.5)(s-0.5)$   \\
      degrees of freedom & 174 \\
      root mean square of residuals & 0.000270536 \\
      \hline
    \end{tabular}
    \begin{tabular}{c|c|cc}
      parameter & value & asymptotic standard error & \\
      \hline
$c_2$               & 0.00426427   &    +/- 8.578e-05  &  (2.012\%)\\
      \hline
    \end{tabular}
  \end{center}
  \caption{\label{tab:fit:locusts}Fitting data, swarm alignment in locusts, Fig.~\ref{fig:densityEst:yates}.}
\end{table}